\documentclass[final,5p,times,twocolumn]{elsarticle}

\usepackage{amssymb}
\usepackage{lineno}
\modulolinenumbers[5]

\usepackage{amsmath, amsfonts}
\usepackage{graphicx}
\usepackage{multirow}
\usepackage{scalerel}
\usepackage{stmaryrd}
\usepackage{booktabs}

\def\x{{\mathbf x}}

\def\B{{\cal B}}

\def\D{{\cal D}}

\def\T{{\cal T}}

\def\E{{\mathbb E}}

\def\MC{{\scaleto{MC}{3pt}}}

\def\SYNSIG{{{SYNSIG}}}
\def\GTSRB{{{GTSRB}}}
\def\Digit{{{Digit}}}
\def\Sign{{{Sign}}}
\def\Visda{{{VisDA}}}

\def\OfficeHome{{{Office-Home}}}

\journal{}
\date{}

\begin{document}

\begin{frontmatter}

\title{Feature Alignment by Uncertainty and Self-Training\\
for Source-Free Unsupervised Domain Adaptation}

\author[sds,seoultech]{JoonHo Lee}

\author[seoultech]{Gyemin Lee}

\address[sds]{Machine Learning Research Center, Samsung SDS Technology Research, Republic of Korea}

\address[seoultech]{Department of Electronic and IT Media Engineering, Seoul National University of Science and Technology, Republic of Korea}

\begin{abstract}
Most unsupervised domain adaptation (UDA) methods assume that labeled source images are available during model adaptation. 
However, this assumption is often infeasible owing to confidentiality issues or memory constraints on mobile devices. 
Some recently developed approaches do not require source images during adaptation, but they show limited performance on perturbed images.
To address these problems, we propose a novel source-free UDA method that uses only a pre-trained source model and unlabeled target images.
Our method captures the aleatoric uncertainty by incorporating data augmentation and trains the feature generator with two consistency objectives. 
The feature generator is encouraged to learn consistent visual features away from the decision boundaries of the head classifier.
Thus, the adapted model becomes more robust to image perturbations.
Inspired by self-supervised learning, our method promotes inter-space alignment between the prediction space and the feature space while incorporating intra-space consistency within the feature space to reduce the domain gap between the source and target domains. 
We also consider epistemic uncertainty to boost the model adaptation performance.
Extensive experiments on popular UDA benchmark datasets demonstrate that the proposed source-free method is comparable or even superior to vanilla UDA methods. 
Moreover, the adapted models show more robust results when input images are perturbed. 
\end{abstract}


\end{frontmatter}

% \linenumbers

\section{Introduction}
\label{sec:intro}

Deep neural networks have enabled major breakthroughs on various visual recognition tasks.
Much of their success is attributable to large well-annotated data.
However, collecting large amounts of labeled data is time-consuming and costly.
A solution is to transfer knowledge from a label-rich source domain to a label-scarce target domain.
However, this process is hindered by the difference between the source data distribution and the target data distribution.
Domain adaptation methods are proposed to tackle this {\sl domain shift} problem.

Unsupervised domain adaptation (UDA) considers a form of domain adaption where only unlabeled images are available in the target domain. 
Many existing UDA methods attempt to reduce the domain shift by minimizing the difference between the source and the target feature distributions \cite{DAN,DANN,ADDA,SAFN,MDD,SWD}.
These methods assume that labeled images are available in the source domain and take advantage of the source images during model adaptation.

However, this assumption is infeasible in many cases.
For example, transferring medical records containing private information is strictly prohibited.
In many institutions, business-sensitive data remain confidential.
Memory constraints pose another issue, especially considering the increased demand for mobile services integrating on-device DNN applications.
The limited memory of a mobile device does not allow the storage of large-scale source data such as ImageNet data \cite{ImageNet}.

Recently, UDA approaches that do not access source data have been proposed \cite{3C-GAN, SHOT, SDDA, SoFA, SFDA, AAA, ISFDA}.
Though these approaches have reported promising results in UDA, our investigation reveals that their performance degrades sharply when input images are perturbed. 
We provide detailed discussion in Sec \ref{ssec:discussion}.
%------------------------------------------------------------------------
\begin{figure}[t]
  \centering
  \includegraphics[width=0.96\columnwidth]{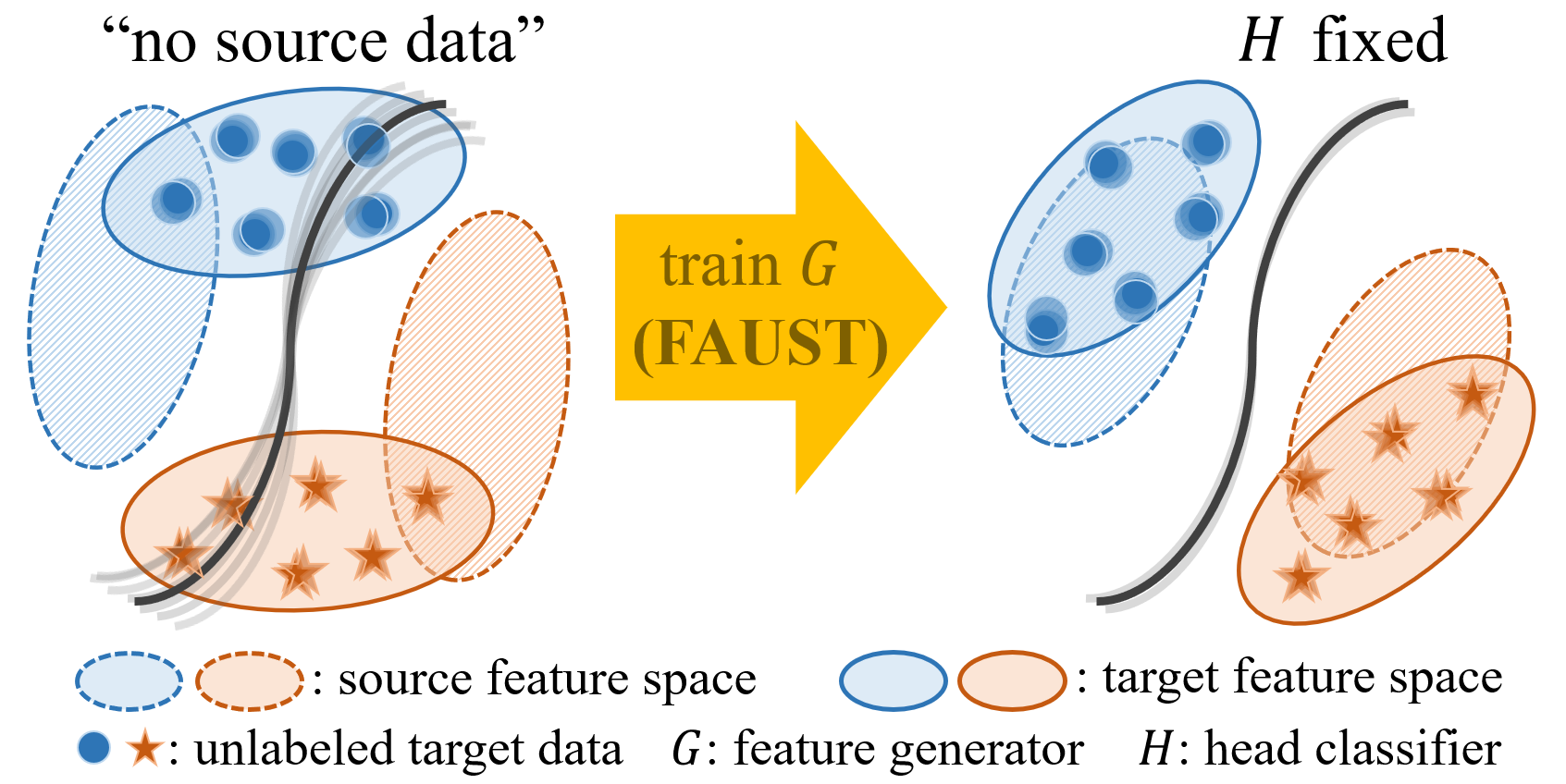}
\caption{
Schematic diagram of the proposed method ({\bf FAUST}). 
To achieve source-free UDA, we fix the head classifier $H$ trained on source images and train the feature generator $G$.
We consider aleatoric (data) and epistemic (model) uncertainties to push target visual features away from the decision boundary of $H$.
Shading around each sample represents aleatoric uncertainty and shading near the decision boundary represents epistemic uncertainty.
}
\label{fig:concept}
\end{figure}
%------------------------------------------------------------------------

To address such limitations, we propose a novel {\sl source-free} UDA method referred to as Feature Alignment by Uncertainty and Self-Training ({\bf FAUST}). 
We assume that only a source model is available, whereas source images are unavailable.
As illustrated in Fig. \ref{fig:concept}, the proposed method freezes the head classifier from the source model and adapts the feature generator to the target images. 
Our key idea is to consider aleatoric uncertainty while training the feature generator to align the features between both domains. 
Aleatoric uncertainty is characterized by noise inherent in images \cite{Aleatory}. 
By considering the effects of noisy images, our feature encoder learns to generate visual features away from the class decision boundaries. 
Thus, the decision boundaries effectively reside in the low-density regions in the feature space.

To capture the aleatoric uncertainty, we propose two consistency objectives by incorporating multiple perturbed views of the same image.
Motivated by self-supervised learning methods \cite{DeepCluster,SwAV}, we constrain different views of the same image to have similar embeddings by imposing {\sl intra-space consistency} in the feature space.
However, the domain gap can still cause incorrect predictions of these consistent images. Thus, we also enforce {\sl inter-space consistency} between the feature space and the prediction space. 
For this purpose, we present a feature-based pseudo-labeling strategy
that selectively uses source-similar target image features. 
As the head classifier from the source model encodes the source distribution, we leverage this information to identify source-similar target images. 
By pseudo-labeling a target image according to the source distribution and by comparing this pseudo-label against predictions from multiple views of the same image, our inter-space consistency encourages the target features to align with the source feature distribution.

We also consider epistemic uncertainty for source-free UDA.
Whereas aleatoric uncertainty is more effective, epistemic uncertainty has proven to be useful for UDA \cite{MUDA,DTA}. 
Along with the proposed two consistency losses capturing aleatoric uncertainty, 
FAUST incorporates the epistemic uncertainty loss based on Monte Carlo dropout sampling.

Our extensive experiments and analysis confirm that FAUST is comparable or superior to recent UDA methods, even completely without the use of source images. 
Furthermore, the adapted models using the proposed method show more robust performance under target image perturbations.

%==========================================================================
\section{Related Work}
\label{sec:related_work}

\noindent
\textbf{Unsupervised Domain Adaptation.} 
Given labeled source images and unlabeled target images, the objective of UDA is to address the domain shift between the source domain and the target domain.
Earlier works align the feature distributions from both domains either by matching their statistical moments \cite{DAN,MDD,SAFN} or by deploying a domain discriminator in an adversarial manner \cite{DANN,ADDA,MDD,SWD,CADA}. 
Some recent works use stochastic predictions to make the model target-discriminative \cite{MCD,DIRT-T,GPDA,STAR,DTA,MUDA} or leverage the optimal transport principle \cite{DeepJDOT,RWOT}.
Semantic augmentation has also been used \cite{TSA}.
Early stopping and block skipping are shown to reduce computing cost \cite{FDAN}.
These methods assume that source images are available during the knowledge transfer process.
This assumption is not valid when the source data is private or confidential. 
This issue creates more difficult source-free problems.

\smallskip \noindent
\textbf{Source-free UDA.} 
Lately, several methods have investigated source-free UDA. 
3C-GAN \cite{3C-GAN} produces target-style samples using a GAN, which collaborates with the source model during adaptation.
SHOT \cite{SHOT} freezes the source classifier and trains the feature encoder by means of pseudo-labeling and mutual information maximization.
SDDA \cite{SDDA} models a GAN combined with a gradient reversal layer to produce domain-invariant features. 
SoFA \cite{SoFA} induces a reference distribution using the source model to extract class semantic features.
SFDA \cite{SFDA} uses pseudo-labels based on low-entropy samples and the point-to-set distance.
AAA \cite{AAA} generates a variety of smooth adversarial examples to generalize the source model to the target distribution.
ISFDA \cite{ISFDA} complements SHOT \cite{SHOT} to overcome class-imbalance scenarios.

\smallskip \noindent
\textbf{Self-Supervised Learning.}
The goal of the proposed method is to learn effective visual features without human annotated labels.
Typical self-supervised methods design pretext tasks to provide automated supervision, where both the inputs and labels are derived from an unlabeled dataset \cite{Color,Jigsaw,Rotation}. 
Many recent contrastive methods build upon the contrastive loss and a set of image transforms \cite{SimCLR,MoCo,BYOL}.
They learn to map different views of the same image nearby and views from different images apart.
Clustering-based methods \cite{DeepCluster} iteratively learn features by clustering images and predicting their cluster assignments. 
SwAV \cite{SwAV} incorporates clustering into contrastive learning by comparing the assignment from one view and predicting it from another view. 
Our pseudo-labeling strategy in the inter-space consistency loss is inspired in part by \cite{SwAV}.
However, rather than using simple $k$-means clustering, our approach uses confident features.

\smallskip \noindent
\textbf{Data Augmentation}
In many image recognition tasks, data augmentation is known to be effective to encode the feature invariance \cite{image_aug}. 
Simple transformations such as horizontal flipping, random cropping and random rotation are widely used in practice.
Recently, more sophisticated image augmentation methods that combine multiple transformations have been proposed \cite{AutoAugment, RandAugment}
to achieve state-of-the-art recognition performance \cite{FixMatch, MPL}. 
When labeled data is available, AutoAugment \cite{AutoAugment} applies reinforcement learning to learn the augmentation strategy. On the other hand, RandAugment \cite{RandAugment} does not require labeled data. Instead, RandAugment randomly select transformations for each image.
Because labeled images are not available in our source-free setting, we incorporate RandAugment to our UDA method to capture the aleatoric uncertainty.

\smallskip \noindent
\textbf{Uncertainty.} 
The predictive uncertainties are generally grouped as epistemic or aleatoric \cite{Uncertainty}. 
Epistemic uncertainty, known as model uncertainty, can be estimated by means of Monte Carlo dropout sampling \cite{Gal16}.
Among UDA methods, \cite{MUDA,DTA} considered epistemic uncertainty. 
Aleatoric uncertainty is described as the noise inherent in images and is known to be more effective in computer vision. 
It can be captured by a distribution over the model outputs. 
Alternatively, test-time data augmentation can be used to estimate the aleatoric uncertainty \cite{TTA}.

%==========================================================================
\section{Preliminary and Notations}
\label{sec:preliminary}

The proposed method addresses the UDA problem in which only a pre-trained source model is available, whereas access to the source data is prohibited.
Conventional UDA methods assume that we are given a set of fully-labeled images $(X_s, Y_s)$ from the source domain $\D_s$ and a set of unlabeled images $X_t$ from the target domain $\D_t$.
However, in more challenging source-free UDA problems, the source images $(X_s, Y_s)$ are no longer available for model adaptation.
Instead, we are given a source model $F_s$ trained on $(X_s, Y_s)$.
In this paper, we use only $F_s$ and $X_t$ to build a model $F_t$ adapted to the target domain $\D_t$.

We consider $K$-way classification where the source domain and the target domain share the same label space. 
In our formulation, a source classification model is divided into two parts: a feature generator $G$ and a task-specific head classifier $H$. 
We fix the head classifier $H$ to focus on adapting the feature generator $G$ to the target domain. 
A decent source model holds a considerable amount of information about the source data. 
In particular, the feature distribution of source embeddings is encoded in the head classifier. 
Therefore, preserving the head classifier and making full use of it is a reasonable strategy in the source-free setting.

We let $F = H \circ G$ denote a model with a feature generator $G$ and head classifier $H$.
Given an image $x$, its feature embedding is $z = G(x)$.
Its $K$-dimensional prediction output is $p = p(x) = \sigma(H(G(x)))$, where $\sigma$ is a softmax function.

%------------------------------------------------------------------------
\begin{figure*}[!t]
  \centering
  \includegraphics[width=0.95\textwidth]{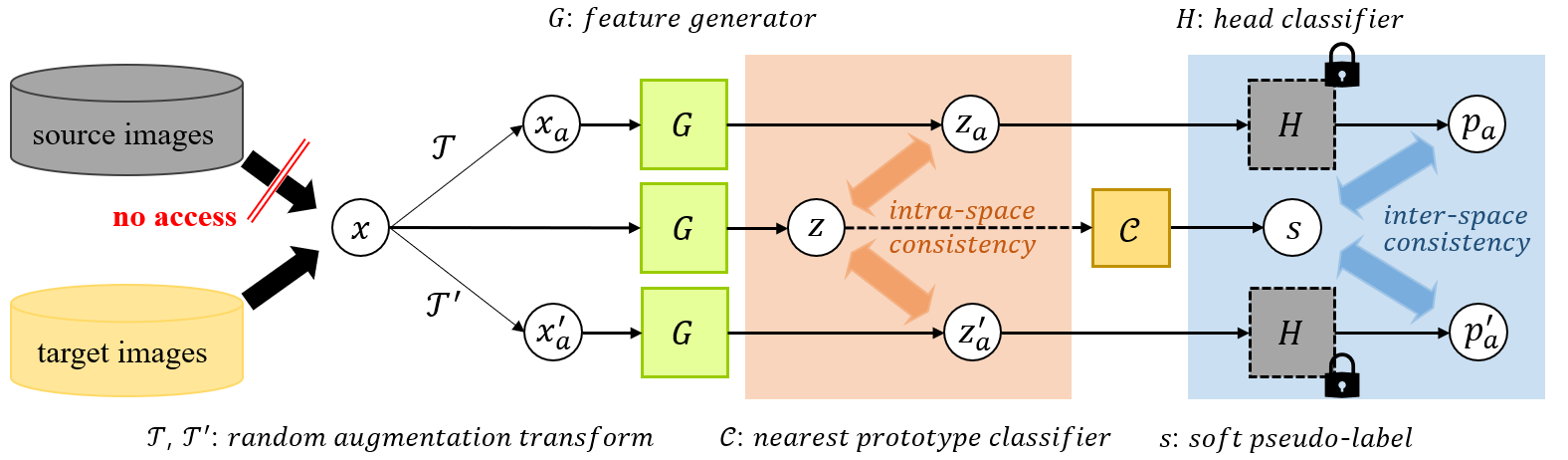}
\caption{
Framework of the proposed Feature Alignment by Uncertainty and Self-Training ({\bf FAUST}).
FAUST aims to learn consistent visual features by considering predictive uncertainties.
When considering aleatoric (data) uncertainty, FAUST incorporates data augmentations ($v$=2) and enforces intra-space consistency in the feature space and inter-space consistency between the feature space and the prediction space.
Generated target features are encouraged to be away from the decision boundaries of the frozen head classifier $H$.
}
\label{fig:architecture}
\end{figure*}
%------------------------------------------------------------------------

%=====================================================================
\section{Proposed Method}

Our source-free UDA considers aleatoric uncertainty to learn consistent target visual features. 
To capture the aleatoric uncertainty, we incorporate data augmentation transforms and promote consistency within the feature space.
Our feature generator is encouraged to learn features that are away from the decision boundaries of the fixed head classifier. 
By leveraging the head classifier that encodes the source feature distribution, the inter-space consistency across the feature space and the prediction space mitigates the domain gap.
We also model epistemic uncertainty for an additional performance gain.

\subsection{Aleatoric Uncertainty by Augmentation}
\label{ssec:uncertainty}

We propose to consider aleatoric uncertainty for UDA tasks. 
Aleatoric uncertainty describes noise inherent in images. 
Our intuition is that by considering the effects of noisy target images, we can encourage the feature extractor to generate features away from the decision boundaries of the head classifier. 
One can envision that images with higher uncertainty are likely to be located around the decision boundaries. 
Hence, even small perturbations can lead to different decision outputs. 
On the other hand, images with higher confidence (lower uncertainty) will be distant from the decision boundaries, and their outputs will remain consistent.
Thus, we aim to train the feature generator $G$ to generate consistent features under perturbations, 
while leaving the head classifier $H$ fixed.

To capture the aleatoric uncertainty, we incorporate data augmentation into our UDA method. 
Given an image $x$, let $x_a^{(1)}, \ldots, x_a^{(v)}$ be a set of $v$ views under a random augmentation transform $\T$. 
As described, these transformed views of the same image should have similar feature representations. 
This insight motivates our {\bf intra-space consistency} objective. 
To make their feature embeddings located more closely to the feature embedding of $x$, we minimize their distances 
\begin{align}
	{\cal L}_{f} = \frac{1}{v} \sum_{i=1}^{v}D(z, z_a^{(i)})
\end{align}
where $D$ denotes the cosine dissimilarity between two feature embeddings 
$z=G(x)$ and $z_a^{(i)}=G(x_a^{(i)})$.

%=====================================================================
\subsection{Feature Alignment by Self-Training}
\label{ssec:FAUST}

We promote consistency in the feature space. 
However, intra-space consistency is not sufficient for UDA tasks because some ``consistent'' target images can be incorrectly predicted due to a domain shift. 
To reduce the domain gap between the source domain $\D_s$ and the target domain $\D_t$, we also propose {\bf inter-space consistency} across the feature space and the prediction space. 
Our inter-space consistency objective encourages the target features to align with the source feature distribution.

For our purpose, the frozen head classifier $H$ plays a key role. 
In a head classifier trained on source images, ample information about the source distribution in the feature space and the decision boundaries for class prediction is distilled. 
We leverage this information to enforce consistency between the pseudo-labels generated in the feature space and the classification outputs in the prediction space.

We compute the pseudo-labels in the feature space by matching the feature embeddings to a set of prototype vectors in the feature space. 
If the source feature distribution is encoded in these prototypes, the target images will be labeled according to the source distribution. 
Ideally, these prototypes can be obtained from the source feature embeddings. 
However, this is not possible in our source-free UDA setting. 
Instead, we use the target images and the head classifier to estimate these prototypes. 
One can easily speculate that target images similar to the source images will be classified with high confidence. 
Hence, we propose to use source-similar target images to produce prototype vectors. 
The detailed procedure for computing prototype vectors and pseudo-labeling is presented below.

Once we obtain the pseudo-labels, we compare the pseudo-label of an image against the prediction outputs of multiple augmented versions of the same image. 
Let $s$ denote the pseudo-label of target image $x$. 
The proposed {inter-space consistency} loss is formulated as follows:
\begin{align}
   {\cal L}_{i} = \frac{1}{v} \sum_{i=1}^{v} {\cal H} (s, p_a^{(i)}).
\end{align}  
Here, $p_a^{(i)}$ is the prediction output of random-augmented view $x_a^{(i)}$, 
and ${\cal H}(s,p_a^{(i)})$ denotes the cross-entropy between $s$ and $p_a^{(i)}$. 
Therefore, our method enforces consistency between the feature space and the prediction space. 
This loss also ensures different augmentations of the same image to have similar feature embeddings and prediction outputs.

To summarize, FAUST aims to learn features away from the decision boundaries by leveraging aleatoric uncertainty.
Given an image, its neighbors (noisy versions obtained by augmentation) must be mapped closely in the feature space (${\cal L}_f$) and must have consistent prediction outputs (${\cal L}_i$).
As a result, the feature of the image will maintain its distance from the decision boundaries. 
Thus, modeling aleatoric uncertainty in this way will effectively locate the boundaries in the low density regions in the feature space.
Fig. \ref{fig:architecture} illustrates our intra-space and inter-space consistency approach.

\smallskip \noindent
\textbf{Pseudo-Labeling with Confidence-Weighted Prototypes.}
\noindent
Here, we describe our pseudo-labeling strategy. 
We pseudo-label feature space vectors using a soft nearest prototype classifier. 
We compute each prototype $c_k$ by evaluating
\begin{align}
	c_k = \sum_{j=1}^{|\B|} p_{jk} z_j
	\label{eq:prototype}
\end{align}
where $z_j$ is the feature embedding of the non-augmented target image $x_j$ in mini-batch $\B$, 
and $p_{jk}$ is the prediction probability of $x_j$ to class $k$.
We assign class conditional confidence $p_{jk}$ as weights to each feature vector.
Because target images similar to source images will have higher confidence and dissimilar ones will have lower confidence, the prototype vectors generated using only target images will be similar to the prototypes from the source images. 

To obtain the soft pseudo-label of a target image $x$, we match its feature $z$ to the set of $K$ prototype vectors. 
Let $C$ denote the matrix whose columns are the prototypes $c_1, \ldots c_K$.
The matching is done by the cosine similarity
\begin{align}
	s = \widehat{\sigma} ( C^T z )
\end{align}
where the feature vector $z$ and each column vector (prototype) of $C$ are normalized by division by their L2-norms during the similarity computation.
Here, $\widehat{\sigma}$ denotes a softmax function with a sharpening temperature. 

Due to the domain difference, a target image can be incorrectly predicted with high confidence by the source classifier. 
However, this incorrect prediction does not immediately lead to incorrect pseudo-labels because multiple images are accumulated to compute the prototypes in equation \eqref{eq:prototype}.
Misclassifying most of the target images with high confidence is also less likely to happen. 
In many realistic UDA problems, we expect that there are source-trained classifiers that perform reasonably well on target images \cite{Ben10}.

%=====================================================================
\subsection{Epistemic Uncertainty and Entropy}
\label{ssec:epistemic}

\noindent
\textbf{Epistemic Uncertainty.} 
Our intra-space and inter-space consistency losses are designed to capture aleatoric uncertainty. 
We can also account for epistemic uncertainty for UDA tasks. 
Epistemic uncertainty explains model uncertainty due to a lack of data and can be reduced by collecting more data.
Though modeling aleatoric uncertainty is more effective in computer vision, epistemic uncertainty has proven to be useful in situations where the training set is small or the test-time (target) images are different from the train-time (source) images \cite{MUDA,DTA}.

We estimate epistemic uncertainty using Monte Carlo (MC) dropout sampling \cite{Gal16}. 
After activating dropout in the model, we perform multiple stochastic forward passes. 
The prediction outputs $p_{\MC}(x)$ of $x$ are called MC dropout samples.
We use the L2-norm of their sample standard deviation to compute the {\bf epistemic uncertainty} loss
\begin{align}
	{\cal L}_{u} = \E_x [\| Std(p_{\scaleto{MC}{3pt}}(x)) \|]
\end{align}
where $Std$ denotes the sample standard deviation.

\smallskip \noindent
\textbf{Entropy Minimization.} 
The conditional entropy is known to be very effective when used to capture cluster assumptions \cite{EntMin} and has been adopted by many UDA methods \cite{SHOT, DTA, Advent}.
Under the cluster assumption, decision boundaries away from high-density regions are preferred.
Because this assumption conforms to our consistency losses, we minimize the conditional {\bf entropy}
\begin{align}
	{\cal L}_{e} = {\cal H}(p) = -\E_{x} [p(x)^T \log p(x)]
	\label{eq:loss_entmin}
\end{align}
to train our feature generator. 
This loss also encourages our feature generator to learn confident (low entropy) features.

%=====================================================================
\subsection{Overall Objective and Optimization}
\label{ssec:optimization}

Given the source model $F_s = H_s \circ G_s$ and the unlabeled target images $X_t$, we fix $H_s$ and train the target-domain feature generator $G_t$ with the following objective:
\begin{equation}
  \min_{G}~ {\cal L}_{i} + \alpha {\cal L}_{f} + \beta {\cal L}_{e} + \gamma {\cal L}_{u}
  \label{eq:total_loss_general}
\end{equation}
where $\alpha, \beta \geq 0$ are tradeoffs.
Considering the computational burden of MC dropout sampling, we can choose to turn on/off ${\cal L}_u$ by setting $\gamma$ equal to 1 or 0. 
These settings are respectively referred as {\bf FAUST+U} and {\bf FAUST} in Sec. \ref{sec:experiment}.

During training, 
prototype computation and pseudo-labeling steps are performed in each mini-batch.  
Contrary to global approaches \cite{SHOT, DeepCluster}, our in-batch approach incurs a lower computational cost when the size of the target data is enormous or increasing.
This simple end-to-end training strategy as well as no network customization are among the advantages of our method.

%==========================================================================
\section{Experiment}
\label{sec:experiment}

We evaluate the proposed method on popular benchmark datasets with smaller images (Digit and Sign) and larger images (Office-Home and VisDA). 
We also extend our validation to more complex multi-source domain adaptation (MSDA) tasks using miniDomainNet.

%=====================================================================
\subsection{Setup}
\label{ssec:setup}

\noindent
\textbf{Digit.} 
We test our method using the three standard digit recognition datasets of SVHN \cite{SVHN}, MNIST \cite{MNIST} and USPS \cite{USPS}. 
The goal is to classify an image into one of ten digits. 
We consider four UDA tasks: 
SVHN$\rightarrow$MNIST (S$\shortrightarrow$M), 
MNIST$\rightarrow$SVHN (M$\shortrightarrow$S), 
MNIST$\rightarrow$USPS (M$\shortrightarrow$U) and 
USPS$\rightarrow$MNIST (U$\shortrightarrow$M).
We note that M$\shortrightarrow$S is most challenging and is often ignored in literature.

\smallskip
\noindent
\textbf{Sign.} 
We also experiment on two traffic sign datasets: Synthetic Signs (\SYNSIG) \cite{SYNSIG} and the German Traffic Sign Recognition Benchmark (\GTSRB) \cite{GTSRB}, sharing 43 classes. 
{\SYNSIG} contains 100K synthetic traffic sign images and {\GTSRB} has more than 50K actual traffic sign images. 
We evaluate the UDA task of \SYNSIG$\rightarrow$GTSRB (S$\shortrightarrow$G).

\smallskip
\noindent
\textbf{Office-Home.} 
Office-Home \cite{Office-Home} contains 15,500 larger images of 65 categories from four distinct domains: Artistic images (Ar), Clip Art (Cl), Product images (Pr), and Real-World images (Rw). 
We consider all 12 UDA tasks.

\smallskip
\noindent
\textbf{VisDA.} 
VisDA-C \cite{Visda17} is a large-scale benchmark including 152K synthetic 3D model images rendered from different angles and lighting conditions and 55K photo-real images sampled from MSCOCO \cite{MSCOCO}. 
This benchmark is one of the most challenging UDA tasks because it contains a large number of complex images and considers a more practical synthetic-to-real UDA problem.

\smallskip
\noindent
\textbf{MiniDomainNet.} 
We extend our evaluation to MSDA tasks using miniDomainNet \cite{DAEL}.
MiniDomainNet is a subset of DomainNet \cite{M3SDA} containing 140K 96$\times$96 images of 126 categories from four different domains: Clipart (Cl), Painting (Pa), Real (Re) and Sketch (Sk).
Unlike single-source tasks, the goal of MSDA is to adapt multiple source domains to a target domain. 
We evaluate FAUST on four MSDA tasks: ${\cal R}\rightarrow$Cl, ${\cal R}\rightarrow$Pa, ${\cal R}\rightarrow$Re and ${\cal R}\rightarrow$Sk, where ${\cal R}$ denotes the remaining three domains apart from the target domain. 
For example, Pa, Re and Sk are the source domains in ${\cal R}\rightarrow$Cl.
Following a setup of \cite{DAEL}, we hold out 630 test images per domain.

\smallskip
\noindent
\textbf{Baseline Methods.} 
We compare our method with state-of-the-art works for vanilla UDA that require source data: 
DANN \cite{DANN}, ADDA \cite{ADDA}, MCD \cite{MCD}, 
SWD \cite{SWD}, SAFN \cite{SAFN}, DTA \cite{DTA}, 
IEDA \cite{IEDA}, STAR \cite{STAR}, GVB-GD \cite{GVB}, 
MUDA \cite{MUDA}, CDAN \cite{CDAN}, BNM \cite{BNM}, 
MDD \cite{MDD}, RWOT \cite{RWOT} and TSA \cite{TSA}.
Our method is also compared with recent source-free UDA methods:
SDDA \cite{SDDA}, SoFA \cite{SoFA}, SFDA \cite{SFDA}, SHOT \cite{SHOT} and 3C-GAN \cite{3C-GAN}.
For the baseline methods, all results are directly cited from the relevant published papers.

%=====================================================================
\subsection{Implementation Details}
\label{ssec:implement}

\noindent
\textbf{Network Architecture.} 
In all UDA tasks, we use the same network architectures used in prior works \cite{ADDA,MCD} for a fair comparison. 
We use LeNet\cite{MNIST}-variant networks with three convolution layers ($G$) and two linear layers ($H$) for Digit and Sign. 
The input image sizes are set to 28$\times$28 for M$\leftrightarrow$U, 32$\times$32 for S$\leftrightarrow$M, and 40$\times$40 for S$\shortrightarrow$G. 
For Office-Home and VisDA, the input images of 224$\times$224 are standardized using ImageNet statistics before being fed into ResNet-50 and ResNet-101 \cite{ResNet}, respectively.
For miniDomainNet, ResNet-18 is used, as in \cite{DAEL}. 
For all ResNet models, we employ a pre-trained ResNet trunk ($G$) followed by two linear layers with 1,000 neurons ($H$).

\smallskip \noindent
\textbf{Source Model Training.} 
We specify the training details of our source models 
for reproducibility.
To train the source models, we use random augmentation and the cosine learning rate decay. 
The initial learning rate is 10$^{-2}$ for the datasets containing smaller images (Digit and Sign) and 10$^{-3}$ for those containing larger images (Office-Home, VisDA and miniDomainNet). 
To validate the source model, we use a validation set if one is available.
Otherwise, we randomly spare 10\% of the train samples.
For Office-Home and VisDA, we additionally apply label smoothing \cite{LabelSmoothing} for a fair comparison with \cite{SHOT}.
For a multi-source task, we combine all of the source domains ({\it source-combine}) to train a single source model.

\smallskip \noindent
\textbf{Optimization.} 
We optimize with Adam for S$\shortrightarrow$M and U$\shortrightarrow$M and with SGD (momentum 0.9) for all of the other tasks. 
We apply a fixed learning rate of 2.0$\times$10$^{-4}$ and a weight decay of 5.0$\times$10$^{-4}$ for all experiments.
For VisDA and Office-Home, we employ a cosine decay schedule with an initial learning rate of 5.0$\times$10$^{-4}$. 
The sharpening temperature for pseudo-labeling is set to 0.025.

\smallskip \noindent
\textbf{Uncertainty Setting.} 
FAUST embodies random data augmentation transforms to capture the aleatoric uncertainty.
We apply RandAugment \cite{RandAugment} and Cutout \cite{Cutout}.
The number of transformed views $v$ is set to 2. 
We reduce $v$ to 1 when we experiment with FAUST+U on VisDA because the computational cost is high.

When we explore epistemic uncertainty with FAUST+U, we employ MC dropout sampling \cite{Gal16}. 
For Digit and Sign, we incorporate in-between dropout layers into the LeNet-variant networks and set the dropout rate to 0.4 for $H$  and 0.1 for $G$.
For the ResNet models, the dropout rate of $H$ is set to 0.4, whereas the residual blocks are left intact. 
We draw ten MC dropout samples for Digit and Sign and two samples for the other tasks due to the computational burden. 
The additional dropout layers are disabled unless $\gamma$=1 in equation \eqref{eq:total_loss_general}.

\smallskip \noindent
\textbf{Evaluation Protocol.} 
We set the maximum number of training epochs to 200, 200, 100, 10, and 100
and the mini-batch size to 256, 256, 128, 64, and 256 for Digit, Sign, Office-Home, VisDA and miniDomainNet, respectively. 
Training stops early when the loss converges.
The results are fairly insensitive to the mini-batch size if it is large enough compared to the label set size. 
We report the mean accuracy (and the standard deviation if space is available) from three independent runs.
We indicate the best in \textbf{bold} and the second best in \textbf{\textit{bold italic}}.

%=====================================================================
\subsection{Results}
\label{ssec:results}

\noindent
{\bf Digit and Sign.}
First, we evaluate the performance of the proposed method on the Digit and Sign tasks. 
As shown in Table \ref{tab:digitsign}, 
our method is consistently superior to previous works, including source-access methods. 
In particular, FAUST outscores other methods by large margins on M$\rightarrow$S.
The adaptation from black-and-white handwritten digits (MNIST) to colored street-view house numbers (SVHN) is the most challenging task. 
Because this task is often ignored in the literature, Table \ref{tab:digitsign} shows only a few available results.
On this task, the epistemic uncertainty loss in FAUST+U provides a significant improvement from 85.9\% to 91.3\%.

Furthermore, the performance of our method is close to the target supervised accuracies. 
This result is remarkable considering the use of no source data during adaptation. 
In U$\shortrightarrow$M, \cite{3C-GAN} slightly outperforms our method.
Whereas our input image size is 28$\times$28 as in most prior works
\cite{ADDA,MCD,ADR,GPDA,STAR}, \cite{3C-GAN} used a size of 32$\times$32.
This choice involves a network with more parameters, which can affect its performance.

%------------------------------------------------------------------------
\begin{table}[!t]
\centering
\def\arraystretch{1.1}
% \footnotesize
% \small
\caption{Classification accuracy (\%) on {\Digit} and {\Sign}. 
\textbf{+U} indicates that the epistemic uncertainty loss ${\cal L}_u$ is additionally used. 
Best in \textbf{bold} and second best in \textbf{\textit{bold italic}}.
}
\vspace{0.5ex}
\resizebox{0.98\columnwidth}{!}
{%
\begin{tabular}{c l c c c c c}
\toprule
 & ~~Methods & S$\rightarrow$M & M$\rightarrow$S & M$\rightarrow$U & U$\rightarrow$M & S$\rightarrow$G~~ \\
\midrule
 & ~~Source-only & 70.0 & 47.8 & 77.4 & 84.9 & 77.1 \\ 
\parbox[t]{2mm}{\multirow{10}{*}{\rotatebox[origin=c]{90}{~Source access (vanilla)}}}
 & ~~ADDA \cite{ADDA} & 76.0$^{ 1.8}$& - & 90.1$^{ 0.8}$ & 89.4$^{ 0.2}$  & - \\
 & ~~DRCN\cite{DRCN} & 82.0$^{ 0.1}$ & 40.1 & 91.8$^{ 0.1}$ & 73.7$^{ 0.0}$ & - \\
 & ~~MCD \cite{MCD} & 96.2$^{ 0.4}$ & 28.7 & 94.2$^{ 0.7}$ & 94.1$^{ 0.3}$ & 94.4$^{ 0.3}$ \\ 
 & ~~DIRT-T \cite{DIRT-T} & 99.4 & 76.5 & - & -  & {99.5}\\
 & ~~SWD \cite{SWD} & 98.9$^{ 0.1}$ &- &  98.1$^{ 0.1}$ & 97.1$^{ 0.1}$ & 98.6$^{0.3}$ \\
 & ~~IEDA \cite{IEDA} & 98.9 & 78.5 & 95.0 & 97.5 & -\\
 & ~~MUDA \cite{MUDA} & 99.1$^{ 0.4}$ &- &  {\bf \textit{98.5}}$^{ 0.1}$ & {96.7}$^{ 0.4}$ & {98.6}$^{ 0.5}$ \\
 & ~~STAR \cite{STAR} & 98.8$^{ 0.1}$ &- &  97.8$^{ 0.1}$ & {97.7}$^{ 0.1}$ & 95.8$^{0.2}$ \\
 & ~~RWOT \cite{RWOT} & 98.8$^{ 0.1}$ &- &  {\bf \textit{98.5}}$^{ 0.1}$ & {97.5}$^{ 0.2}$ & - \\
 & ~~TSA \cite{TSA} & 98.7$^{0.2}$ & - & 98.0$^{0.1}$ & 98.3$^{0.3}$ & -\\
\midrule
\parbox[t]{2mm}{\multirow{5}{*}{\rotatebox[origin=c]{90}{Source-free}}}
 & ~~SDDA \cite{SDDA} & 76.3 &- &  89.9 & - & - \\
 & ~~SHOT \cite{SHOT} & 98.9$^{ 0.0}$ &- &  98.0$^{ 0.2}$ & 98.4$^{ 0.6}$ & - \\
 & ~~3C-GAN \cite{3C-GAN} & 99.4$^{ 0.1}$ &- &  97.3$^{ 0.2}$ & \textbf{99.3}$^{ 0.1}$ & {\bf \textit{99.6}}$^{0.1}$ \\
 & ~~\textbf{FAUST} ($v$=2)~~ & {\bf 99.6}$^{ 0.0}$ & {\bf \textit{85.9}}$^{0.2}$ & 98.3$^{ 0.1}$ & {98.8}$^{ 0.0}$ & 99.5$^{ 0.1}$ \\
 & ~~\textbf{FAUST+U} ($v$=2)~~ & {\bf 99.6}$^{ 0.0}$ & {\bf 91.3}$^{0.1}$ & {\bf 98.8}$^{ 0.1}$ & {\bf \textit{99.1}}$^{ 0.1}$ & {\bf 99.7}$^{ 0.0}$ \\
\midrule
 & ~~Target Supervised & 99.6$^{ 0.0}$ & 92.5$^{0.3}$ &99.5$^{ 0.1}$ & {99.5}$^{ 0.1}$ & {99.8}$^{0.1}$ \\
\bottomrule
\end{tabular}
}%
\vspace{-1ex}

\label{tab:digitsign}
\end{table}
%------------------------------------------------------------------------

\smallskip \noindent
{\bf Office-Home.}
FAUST shows accuracies comparable to those of the best baseline method, as can be seen in Table \ref{tab:officehome}.
FAUST+U improves FAUST on most Office-Home tasks and records the highest average accuracy of 72.0\%. 
This result shows that the epistemic uncertainty loss $L_u$ in FAUST+U enhances UDA by complementing the aleatoric uncertainty losses.

%------------------------------------------------------------------------
\begin{table*}[!t]
\centering
\def\arraystretch{1.1}
\caption{ Classification accuracy (\%) on {\OfficeHome} (ResNet-50). 
}
\resizebox{0.90\textwidth}{!}
{%
\begin{tabular}{l c c c c c c c c c c c c|c}
\toprule
Methods & Ar$\shortrightarrow$Cl & Ar$\shortrightarrow$Pr & Ar$\shortrightarrow$Re & Cl$\shortrightarrow$Ar & Cl$\shortrightarrow$Pr & Cl$\shortrightarrow$Re & Pr$\shortrightarrow$Ar & Pr$\shortrightarrow$Cl & Pr$\shortrightarrow$Re & Re$\shortrightarrow$Ar & Re$\shortrightarrow$Cl & Re$\shortrightarrow$Pr & Avg.\\ 
\midrule
Source-only &43.2   &57.9   &71.1   &52.2   &61.4   &60.3   &47.4   &39.7   &64.7   &66.5   &50.2   &80.5   &57.9\\ 
DANN \cite{DANN}	&45.6	&59.3	&70.1	&47.0	&58.5	&60.9	&46.1	&43.7	&68.5	&63.2	&51.8	&76.8	&57.6 \\
CDAN \cite{CDAN}	&50.7	&70.6	&76.0	&57.6	&70.0	&70.0	&57.4	&50.9	&77.3	&70.9	&56.7	&61.6	&64.1 \\
SAFN \cite{SAFN}	&52.0	&71.7	&76.3	&64.2	&69.9	&71.9	&63.7	&51.4	&77.1	&70.9	&57.1	&81.5	&67.3 \\
MDD \cite{MDD}	&54.9	&73.7	&77.8	&60.0	&71.4	&71.8	&61.2	&53.6	&78.1	&72.5	&60.2	&82.3	&68.1 \\
% CDAN+BNM
BNM \cite{BNM}	&56.2	&73.7	&79.0	&63.1	&73.6	&74.0	&62.4	&54.8	&80.7	&72.4	&58.9	&83.5	&69.4 \\
GVB-GD \cite{GVB}	&{57.0}	&74.7	&{79.8}	&64.6	&{74.1}	&74.6	&65.2	&55.1	&{81.0}	&{74.6}	&59.7	&84.3	&{70.4} \\
TSA \cite{TSA} &57.6 &75.8 &{80.7} &64.3 &76.3 &75.1 &{\bf \textit{66.7}} &55.7 &81.2 &{\bf 75.7} &61.9 &83.8 &71.2\\
\midrule
SoFA \cite{SoFA} &-  &74.1   &77.6   &-   &71.8  &75.1   &-   &-   &-  &-   &-   &- &-   \\
SFDA \cite{SFDA} &48.4  &73.4   &76.9   &64.3   &69.8  &71.7   &62.7   &45.3   &76.6  &69.8   &50.5   &79.0 & 65.7   \\
SHOT \cite{SHOT} & 57.1 & 78.1 & {\bf \textit{81.5}} & \textbf{68.0} & {\bf \textit{78.2}} & {\it \textbf{78.1}} & {\it \textbf{67.4}} & 54.9 & \textbf{82.2} & 73.3 & 58.8 & 84.3 & {\bf \textit{71.8}} \\ 
AAA \cite{AAA} & 56.7 & 78.3 & \textbf{82.1} & {\bf \textit{66.4}} & {\bf {78.5}} & {\bf 79.4} & {\bf 67.6} & 53.5 & {\it \textbf{81.6}} & 74.5 & 58.4 & 84.1 & {\bf \textit{71.8}} \\
\textbf{FAUST} ($v$=2) & {\bf \textit{59.4}} &	{\bf \textit{78.5}}&	79.4&	62.7&	77.6&	75.0&	64.5&	{\bf{61.0}}&	78.3&	72.7&	{\bf{64.8}}&	{\bf \textit{85.9}}&	71.6\\
\textbf{FAUST+U} ($v$=2) & {\bf {61.4}}&	{\bf 79.2}&	79.6&	63.3&	76.9&	75.2&	{65.3}&	{\bf \textit{59.4}}&	79.0&	{\bf \textit{74.7}}&	{\bf \textit{64.2}}&	{\bf{86.1}}&	{\bf 72.0}\\
\bottomrule
\end{tabular}
}%
\vspace{-1ex}
\label{tab:officehome}
\end{table*}
%------------------------------------------------------------------------

\smallskip \noindent
{\bf VisDA.}
We report the performance of FAUST on the challenging VisDA dataset in Table \ref{tab:visda}. 
For each of the twelve categories, the accuracy is improved by a large margin over the source-only model, by more than 28\% on average. 
Our approach outperforms all of the baselines and establishes a new state-of-the-art average accuracy of 85.2\%.
This result of source-free FAUST is better than that of the best vanilla UDA method that uses the source data.

Table \ref{tab:visda} shows the effect of modeling the epistemic uncertainty. 
The number of transformed views $v$ for FAUST+U is reduced to 1 due to memory constraints. 
For a fair comparison, FAUST with $v$=1 is presented. 
Adding the epistemic uncertainty loss $L_u$ to FAUST slightly raises the accuracy.
Table \ref{tab:visda} also suggests that increasing $v$ from 1 to 2 improves the performance. 
The accuracy is increased by 0.4\% with $v$=2.
We analyze the effects of more views in Sec \ref{ssec:discussion}.

We note that this result on VisDA as well as the result on Sign (S$\shortrightarrow$G) demonstrate that our approach can be effective during difficult adaptation tasks from synthetic images to real images, even in a source-free setting.

%------------------------------------------------------------------------
\begin{table*}[!t]
\centering
\def\arraystretch{1.0}
\caption{Classification accuracy (\%) on {\Visda} (ResNet-101).
}
\resizebox{0.88\textwidth}{!}
{%
\begin{tabular}{l c c c c c c c c c c c c| c}
\toprule
~~Methods ~& plane ~& bcycl ~& bus ~& car ~& horse ~& knife ~& mcycl ~& person ~& plant ~& sktbrd ~& train ~& truck ~& ~~Avg.~~ \\ 
\midrule
~~Source-only ~& 70.0 ~& 60.8 ~& 57.9 ~& 59.6 ~& 82.9 ~& 42.5 ~& 80.3 ~& 34.4 ~& 48.8 ~& 38.3 ~& 84.2 ~& 19.9 ~& 56.6 \\
~~MCD \cite{MCD} ~& 87.0 ~& 60.9 ~& {83.7} ~& {64.0} ~& 88.9 ~& {79.6} ~& 84.7 ~& {76.9} ~& {88.6} ~& 40.3 ~& 83.0 ~& 25.8 ~& 71.9  \\
~~SAFN \cite{SAFN} ~&93.6  ~&61.3   ~&{{84.1}}   ~&70.6   ~&{94.1}  ~&79.0   ~&{91.8}   ~&79.6   ~&89.9  ~&55.6   ~&\textbf{\textit{89.0}}   ~&24.4 ~&76.1\\   
~~SWD \cite{SWD} ~&90.8  ~&82.5   ~&81.7   ~&70.5   ~&91.7  ~&69.5   ~&86.3   ~&77.5   ~&87.4  ~&63.6   ~&85.6   ~&29.2 ~&76.4\\   
~~MUDA \cite{MUDA} ~& {92.2} ~& {79.5} ~& 80.8 ~& {70.2} ~& {91.9} ~& {78.5} ~& {90.8} ~& {81.9} ~& {93.0} ~& {62.5} ~& {88.7} ~& {31.9} ~& {78.5}  \\ 
~~DTA \cite{DTA}	    ~&93.7	~&82.2	~&85.6	~&\textbf{\textit{83.8}}	~&93.0	~&81.0	~&90.7	~&82.1	~&{95.1}	~&78.1	~&86.4	~&32.1	~&81.5 \\
~~STAR \cite{STAR}	~& 95.0	~&84.0	~&{84.6}	~&73.0	~&91.6	~&91.8	~&85.9	~&78.4	~&{94.4}	~&84.7	~&87.0	~&42.2	~&82.7 \\
~~RWOT \cite{RWOT}	~&95.1	~&80.3	~&83.7	~&\textbf{90.0}	~&92.4	~&68.0	~&{\bf \textit{92.5}}	~&82.2	~&87.9	~&78.4	~&\textbf{90.4}	~&{\it \textbf{68.2}}	~&84.0 \\

\midrule
~~SoFA \cite{SoFA} ~&- ~&- ~&- ~&- ~&- ~&- ~&- ~&- ~&- ~&- ~&- ~&- ~&64.6\\
~~SFDA \cite{SFDA} ~&86.9 ~&81.7 ~&84.6 ~&63.9 ~&93.1 ~&91.4 ~&86.6 ~&71.9 ~&84.5 ~&58.2 ~&74.5 ~&42.7 ~&76.7\\
~~3C-GAN \cite{3C-GAN} ~& 94.8 ~& 73.4 ~& 68.8 ~& {74.8} ~& {93.1} ~& 95.4 ~& 88.6 ~& {\bf{84.7}} ~& 89.1 ~& 84.7 ~& 83.5 ~& 48.1 ~& 81.6  \\ 
~~SHOT \cite{SHOT} ~& 94.3 ~& \textbf{88.5} ~& 80.1 ~& 57.3 ~& {93.1} ~& 94.9 ~& 80.7 ~& 80.3 ~& 91.5 ~& 89.1 ~& 86.3 ~& 58.2 ~& 82.9  \\ 

~~AAA \cite{AAA} & 94.4 & {\bf \textit{85.9}} & 74.9 & 60.2 & {\bf 96.0} & 93.5 & 87.8 & 80.8 & 90.2 & {\bf 92.0} & 86.6 & {\bf 68.3} & 84.2  \\

~~\textbf{FAUST} ($v$=1) ~& {95.6} ~& {80.4} ~& {85.2} ~& 76.2 ~& 94.8 ~& {\bf 97.3} ~& {91.5} ~& {\bf \textit{84.0}} ~& 92.2 ~& 87.9 ~& {86.9} ~& {45.0} ~& 84.8 \\
~~\textbf{FAUST} ($v$=2) ~& {\bf 96.7} ~& 77.6 ~& {\bf{87.6}} ~& 73.3 ~& {\bf \textit{95.5}} ~& {95.4} ~& {\bf 92.9} ~& {83.6} ~& {\bf \textit{95.3}} ~& {\bf \textit{89.5}} ~& {87.7} ~& {46.9} ~& {\bf 85.2} \\
~~\textbf{FAUST+U} ($v$=1) ~& {\bf \textit{96.0}} ~& 78.2 ~& {\bf \textit{87.0}} ~& 78.0 ~& {94.6} ~& {\bf \textit{96.3}} ~& {90.7} ~& 83.3 ~& {\bf 96.3} ~& {87.9} ~& {86.4} ~& {45.1} ~& {\bf \textit{84.9}} \\

\bottomrule
\end{tabular}
}%
\vspace{-1ex}
\label{tab:visda}
\end{table*}
%------------------------------------------------------------------------

\smallskip \noindent
{\bf MiniDomainNet.}
In more complex multi-source tasks, FAUST shows promising results, as shown in Table \ref{tab:miniDN}.
For MiniDomainNet, we run the comparison with recent multi-source DA methods,
DCTN \cite{DCTN}, M$^3$SDA \cite{M3SDA}, MME \cite{MME} and DAEL \cite{DAEL}.
Despite the naive source-combine setup, 
FAUST achieves a higher average accuracy of 62.0\% than the best multi-source method and significantly outperforms the other baselines. 
We note that FAUST does not use source images during adaptation, unlike the other methods.

Contrary to the previous single-source tasks, FAUST+U shows degraded performance. 
Simply combining multiple source domains makes it difficult for the model to classify as many as 126 object categories. 
Because the features from different domains are not well-aligned, the model uncertainty can be high near the decision boundary. 
In relation to this, we speculate that the network capacity of ResNet-18 cannot sufficiently reduce ${\cal L}_u$. 
When we replace ResNet-18 with a larger network ResNet-50, the performance degradation issue disappears (see Table \ref{tab:miniDN_R50}).

%------------------------------------------------------------------------
\begin{table}[!t]
\centering
\def\arraystretch{1.1}
\caption{ Classification accuracy (\%) on miniDomainNet (ResNet-18). 
${\cal R}$ denotes the remaining source domains.
}
\resizebox{0.92\columnwidth}{!}
{%
\begin{tabular}{l c c c c |c}
\toprule
~Methods & ${\cal R}\shortrightarrow$Cl ~& ${\cal R}\shortrightarrow$Pa ~& ${\cal R}\shortrightarrow$Re ~& ${\cal R}\shortrightarrow$Sk ~& ~Avg.~\\ 
\midrule
~Source-only & 63.4 ~& 49.9 ~& 61.5 ~& 44.1 ~& 54.8 \\ 
~DANN \cite{DANN} &65.6$^{0.3}$ ~&46.3$^{0.7}$ ~&58.7$^{0.6}$ ~&47.9$^{0.5}$ ~&54.6 \\
~MCD \cite{MCD} &62.9$^{0.7}$ ~&45.8$^{0.5}$ ~&57.6$^{0.3}$ ~&45.9$^{0.7}$ ~&53.0 \\
~DCTN \cite{DCTN} &62.1$^{0.6}$ ~&45.8$^{0.5}$ ~&58.9$^{0.6}$ ~&48.3$^{0.3}$ ~&54.5 \\
~M$^3$SDA \cite{M3SDA} &64.2$^{0.3}$ ~&49.1$^{0.2}$ ~&57.7$^{0.2}$ ~&49.2$^{0.3}$ ~&55.0 \\
~MME \cite{MME} &{\bf \textit{68.1}}$^{0.2}$ ~&47.1$^{0.3}$ ~&63.3$^{0.2}$ ~&43.5$^{0.5}$ ~&55.5 \\
~DAEL \cite{DAEL} &{\bf 70.0}$^{0.5}$ ~&{\bf 55.1}$^{0.8}$ ~&66.1$^{0.1}$ ~&55.7$^{0.8}$ ~&{\bf \textit{61.7}} \\
\midrule
~\textbf{FAUST} ($v$=2) &  {\bf \textit{68.1}}$^{0.5}$&	{\bf \textit{52.2}}$^{0.7}$&	{\bf 68.7}$^{0.2}$&	{\bf 59.1}$^{0.6}$&	{\bf{62.0}}\\ 
~\textbf{FAUST+U} ($v$=2) &  67.0$^{0.1}$&	51.9$^{0.4}$&	{\bf \textit{67.1}}$^{0.4}$&	{\bf \textit{57.5}}$^{0.5}$&	60.9\\ 
\midrule
~Target Supervised & 72.6$^{0.3}$ ~& 60.5$^{0.7}$ ~& 80.5$^{0.3}$ ~& 63.4$^{0.2}$ ~& 69.3 \\ 

\bottomrule
\end{tabular}
}%
\vspace{1ex}
\label{tab:miniDN}
\end{table}
%------------------------------------------------------------------------

%------------------------------------------------------------------------
\begin{table}[!t]
\centering
\def\arraystretch{1.1}
\caption{ MiniDomainNet results with ResNet-50. 
}
%\vspace{-1.4ex}
\resizebox{0.92\columnwidth}{!}
{%
\begin{tabular}{l c c c c |c}
\toprule
~Methods & ${\cal R}\shortrightarrow$Cl ~& ${\cal R}\shortrightarrow$Pa ~& ${\cal R}\shortrightarrow$Re ~& ${\cal R}\shortrightarrow$Sk ~& ~Avg.~\\ 
\midrule
~Source-only &68.9 &57.5 &66.4 &56.4 & 62.3 \\
~\textbf{FAUST} ($v$=2) &  74.9$^{0.3}$&	{60.1}$^{0.6}$&	75.0$^{0.8}$&	{\bf 64.6}$^{1.1}$&	{68.6}\\ 
~\textbf{FAUST+U} ($v$=2) &  {\bf 75.0}$^{0.8}$&	{\bf 60.7}$^{0.7}$&	{\bf 75.4}$^{0.8}$&	64.0$^{1.0}$&	{\bf 68.8}\\ 
\midrule
~Target Supervised & 80.8$^{0.5}$ ~& 68.3$^{0.8}$ ~& 85.0$^{0.4}$ ~& 71.0$^{0.3}$ ~& 76.3 \\ 

\bottomrule
\end{tabular}
}%
\vspace{1ex}
\label{tab:miniDN_R50}
\end{table}
%------------------------------------------------------------------------

%=====================================================================
\subsection{Analysis and Discussion}
\label{ssec:discussion}

\smallskip \noindent
\textbf{Inferences on Perturbed Target Images.} 
We investigate the performance of adapted models when the target images are perturbed. 
We compare FAUST+U with a well-known source-free method SHOT-IM \cite{SHOT}.
As shown in Table \ref{tab:perturb_inf}, images in Office-Home dataset are perturbed with different levels of severity. 
With no perturbation, FAUST+U and SHOT-IM differ 1.5\% in accuracy. 
However, SHOT-IM exhibits drastic performance drops (9.2\% and 13.5\%) under target image perturbations. 
On the other hand, performance drops of FAUST+U are limited within 3.1\%.
Thus, the difference between FAUST+U and SHOT-IM becomes significant as large as 12\%.
Because our method considers uncertainty during adaptation, the adapted models show more robust results under image perturbations.
%------------------------------------------------------------------------
\begin{table*}[!t]
\centering
\def\arraystretch{1.1}
\caption{ Accuracy (\%) on Office-Home dataset with different levels of perturbation. 
The performance of SHOT-IM drops sharply with more image perturbation. 
However, FAUST+U shows little performance drops. 
}
\resizebox{0.96\textwidth}{!}
{%
\begin{tabular}{l c c c c c c c c c c c c|c}
\hline
Perturbation & Ar$\shortrightarrow$Cl & Ar$\shortrightarrow$Pr & Ar$\shortrightarrow$Re & Cl$\shortrightarrow$Ar & Cl$\shortrightarrow$Pr & Cl$\shortrightarrow$Re & Pr$\shortrightarrow$Ar & Pr$\shortrightarrow$Cl & Pr$\shortrightarrow$Re & Re$\shortrightarrow$Ar & Re$\shortrightarrow$Cl & Re$\shortrightarrow$Pr & Avg.\\ 
\hline
\multicolumn{13}{l|}{\underline{\bf SHOT-IM}} &  \\
No perturbation & 55.4 &	76.6&	80.4&	66.9&	74.3&	75.4&	65.76&	54.8&	80.7&	73.7&	58.4&	83.4&	70.5 \\
Rotate-Crop-Flip & 50.9	& 67.5	& 71.1	& 60.8	& 65.9	& 63.3	& 57.1	& 47.3	& 68.6	& 61.6	& 48.9	& 72.2	& 61.3 \\
RandAugment \cite{RandAugment} & 45.9 & 62.6 & 67.0 & 54.7 & 64.1 & 60.8 & 55.1 & 42.5 & 61.2 & 58.0 & 43.8 & 68.2 & 57.0 \\
\hline
\multicolumn{13}{l|}{\underline{\bf FAUST+U ($v$=2)}} &  \\
No perturbation & 61.4&	79.2&	79.6&	63.3&	76.9&	75.2&	65.3&	59.4&	79.0&	74.7&	64.2&	86.1&	72.0 \\
Rotate-Crop-Flip & 56.9	& 77.8	& 77.1	& 62.9	& 77.1	& 73.9	& 63.3	& 58.8	& 75.5	& 71.4	& 63.8	& 83.9	& 70.2 \\
RandAugment \cite{RandAugment} & 55.1 & 77.1 & 76.4 & 61.2 & 76.5 & 73.2 & 60.8 & 57.5 & 72.9 & 71.0 & 63.7 & 81.8 & 68.9 \\
\hline
\end{tabular}
}%
\vspace{-1ex}
\label{tab:perturb_inf}
\end{table*}
%------------------------------------------------------------------------

\begin{figure}[!t]
  \centering
  \begin{minipage}[t]{0.46\columnwidth} \centering \footnotesize
  \includegraphics[width=\linewidth]{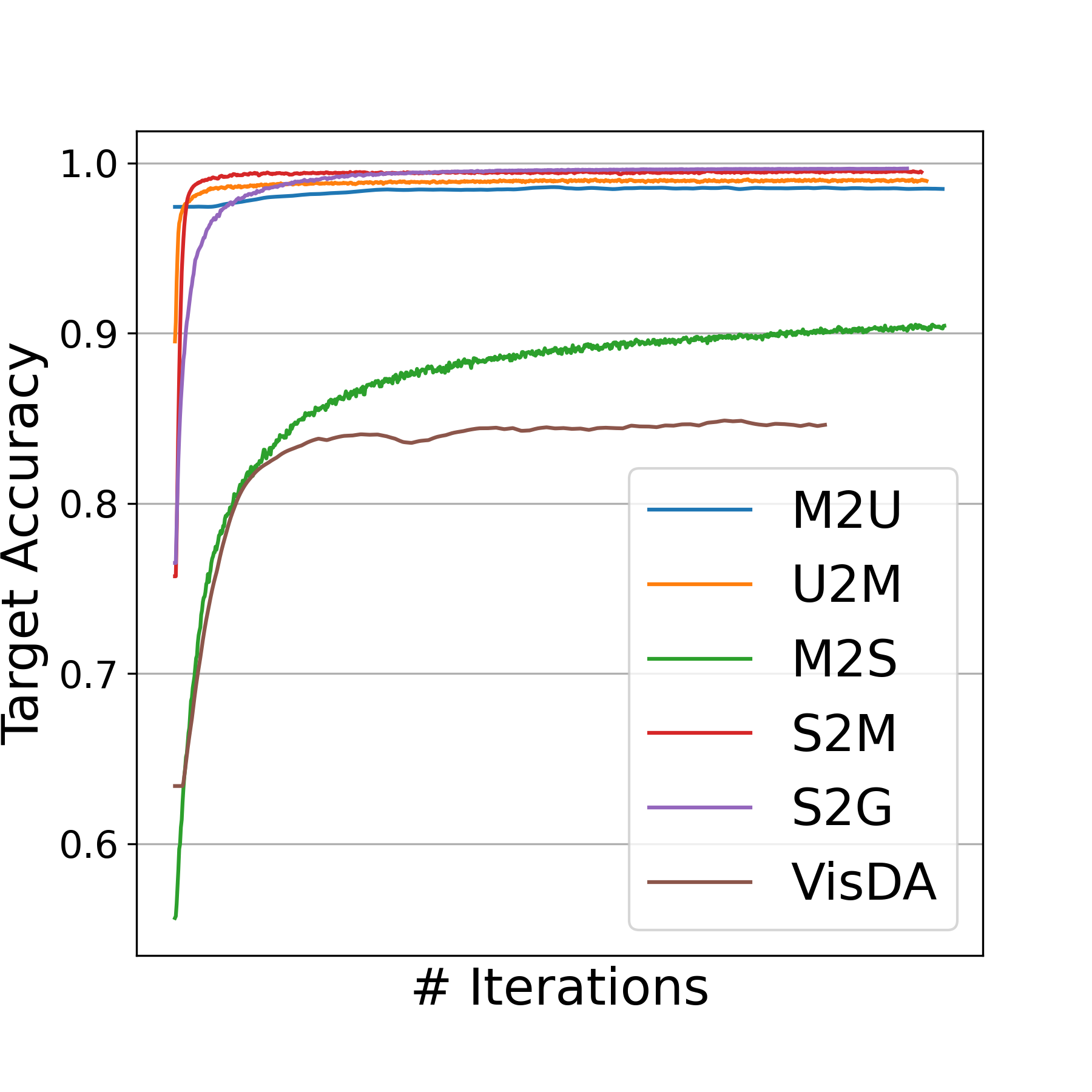}
  (a) Target accuracy
  \end{minipage}
  \begin{minipage}[t]{0.46\columnwidth} \centering \footnotesize
  \includegraphics[width=\linewidth]{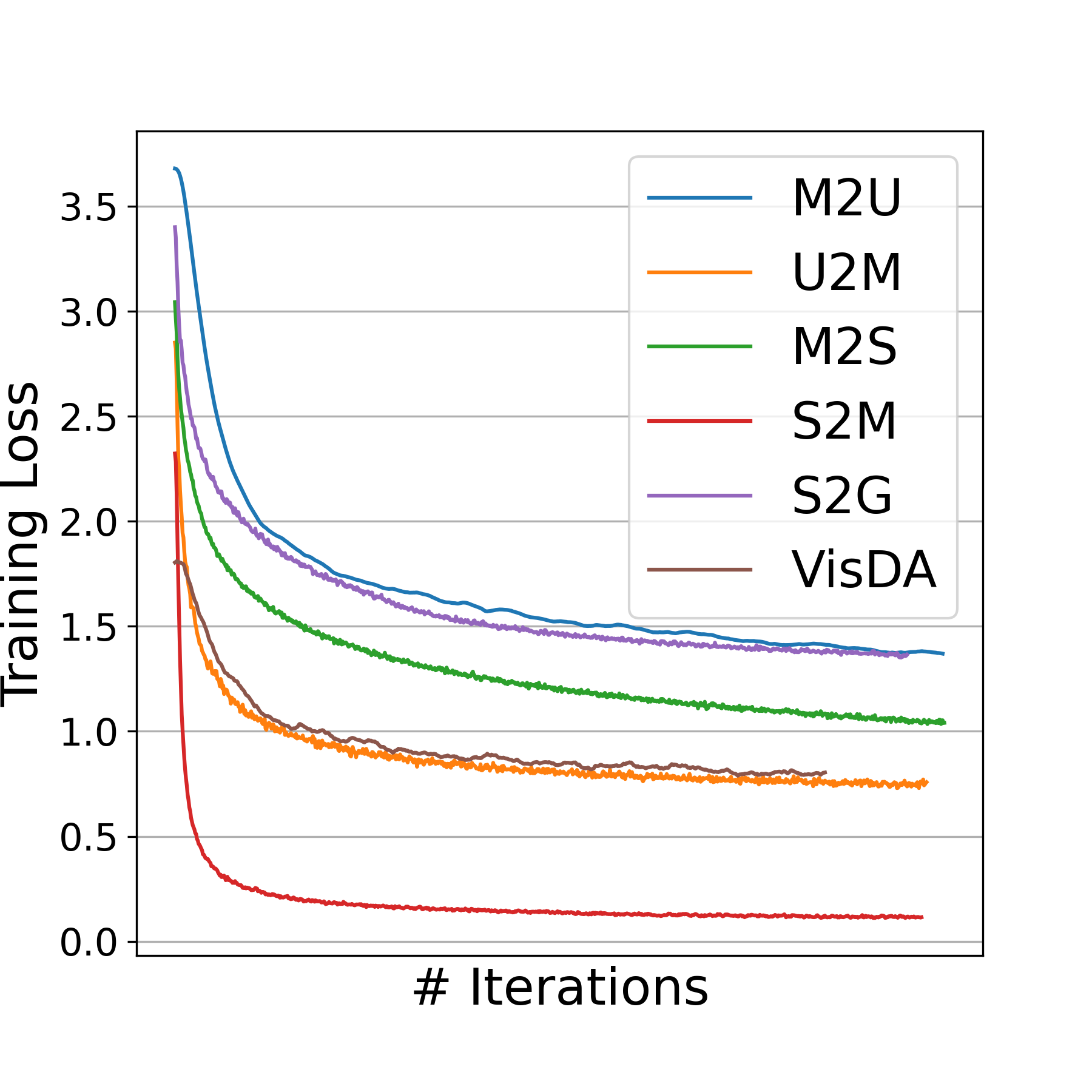}
  (b) Training loss 
  \end{minipage}
\vspace{0.5ex}
\caption{(Best viewed in color) 
The learning curves of (a) target accuracy and (b) training loss on Digit, Sign and VisDA.
The $x$-axis denotes the number of UDA iterations and is rescaled per task.
}
\label{fig:learning_curves}
\end{figure}

\smallskip \noindent
\textbf{Training Stability.}
Learning curves on various benchmark tasks are illustrated in Fig. \ref{fig:learning_curves}.
The horizontal axis of each task is adjusted to a different scale for better visualization. 
The steady change of the target accuracy and training loss over the number of iterations shows that learning is stable during adaptation and converges well.

%------------------------------------------------------------------------
%------------------------------------------------------------------------

\begin{figure}[!t]
  \centering
  \begin{minipage}[t]{0.44\columnwidth} \centering \footnotesize
  \includegraphics[width=0.95\linewidth]{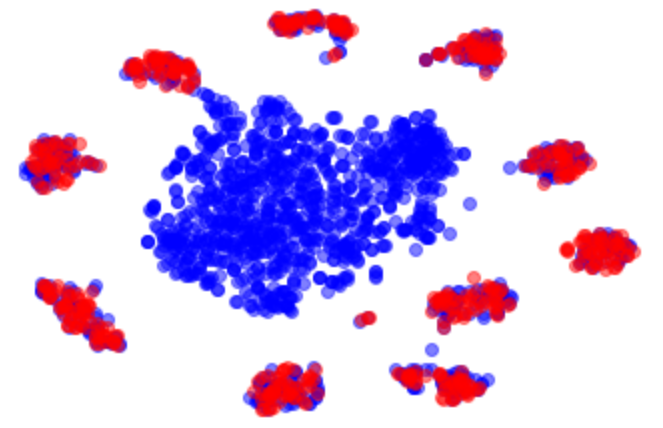}
  (a) Source-only (by domain)
  \end{minipage}~
  \begin{minipage}[t]{0.44\columnwidth} \centering \footnotesize
  \includegraphics[width=0.95\linewidth]{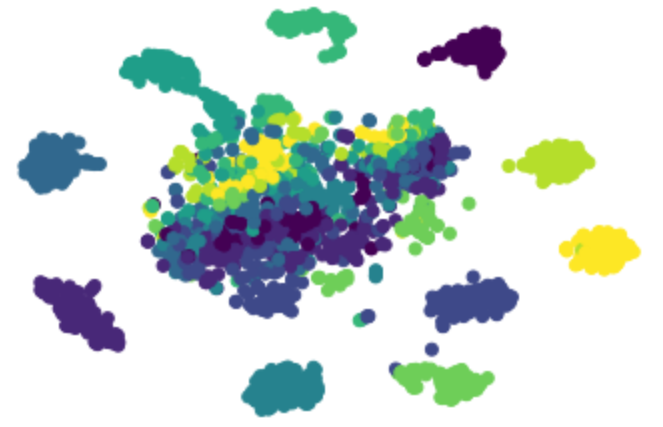}
  (b) Source-only (by class)
  \end{minipage}\\
  \begin{minipage}[t]{0.44\columnwidth} \centering \footnotesize
  \includegraphics[width=0.95\linewidth]{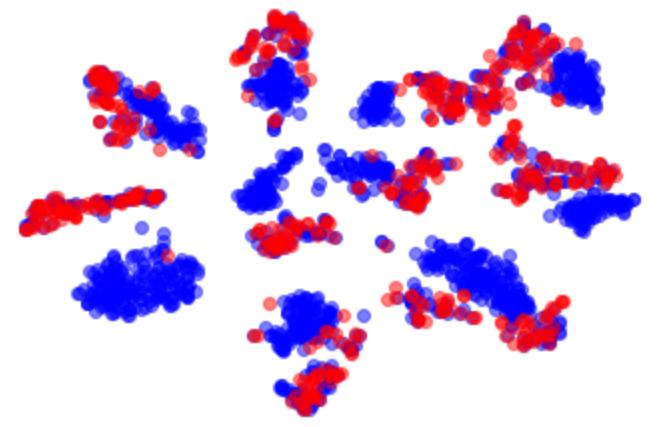}
  (c) Adapted (by domain)
  \end{minipage}~
  \begin{minipage}[t]{0.44\columnwidth} \centering \footnotesize
  \includegraphics[width=0.95\linewidth]{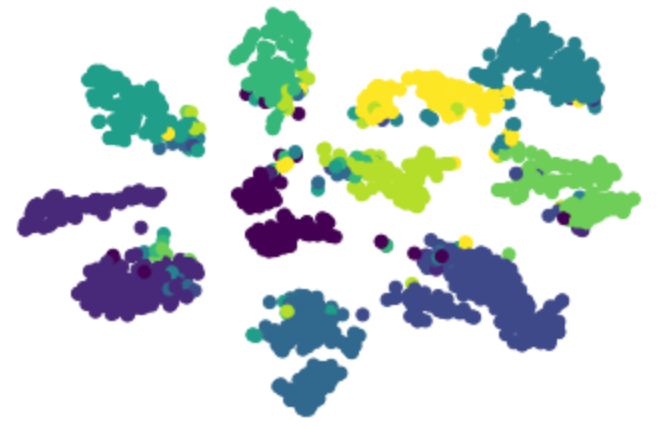}
  (d) Adapted (by class)
  \end{minipage}
\vspace{0.5ex}
\caption{
(Best viewed in color) Feature embeddings from the last pooling layer of $G$ are visualized using t-SNE \cite{Maaten08}. 
{\bf (a) and (c):} Red and blue dots are the test samples from the source (MNIST) and the target (SVHN) domains, respectively.
{\bf (b) and (d):} Each color represents a different class.
}
\label{fig:feature_tsne}
\end{figure}

\begin{figure*}[!t]
  \centering
  \begin{minipage}[t]{0.49\linewidth} \centering \footnotesize
      \begin{minipage}[t]{0.44\linewidth} \centering \footnotesize
      \includegraphics[width=\linewidth]{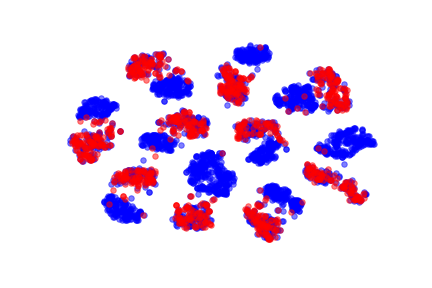}\\
      \vspace{-2.5ex}
      (by domain)
      \end{minipage}
      \begin{minipage}[t]{0.44\linewidth} \centering \footnotesize
      \includegraphics[width=\linewidth]{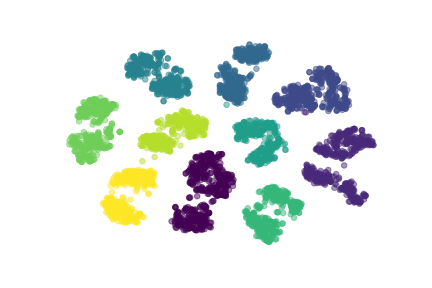}\\
      \vspace{-2.5ex}
      (by class)
      \end{minipage}\\
      \medskip
      \begin{minipage}[t]{0.44\linewidth} \centering \footnotesize
      \includegraphics[width=\linewidth]{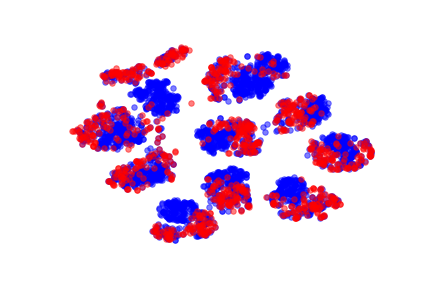}\\
      \vspace{-2.5ex}
      (by domain)
      \end{minipage}
      \begin{minipage}[t]{0.44\linewidth} \centering \footnotesize
      \includegraphics[width=\linewidth]{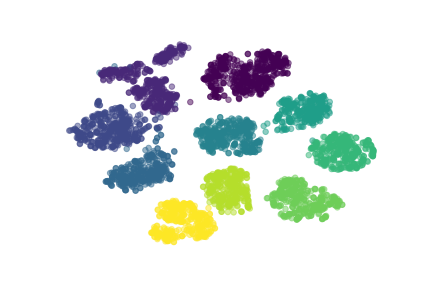}\\
      \vspace{-2.5ex}
      (by class)
      \end{minipage}\\
      \medskip
      (a) M$\rightarrow$U
  \end{minipage}~~
  \begin{minipage}[t]{0.49\linewidth} \centering \footnotesize
      \begin{minipage}[t]{0.44\linewidth} \centering \footnotesize
      \includegraphics[width=\linewidth]{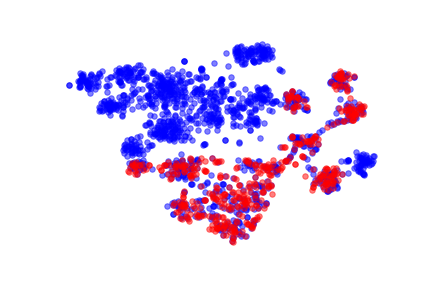}\\
      \vspace{-2.5ex}
      (by domain)
      \end{minipage}
      \begin{minipage}[t]{0.44\linewidth} \centering \footnotesize
      \includegraphics[width=\linewidth]{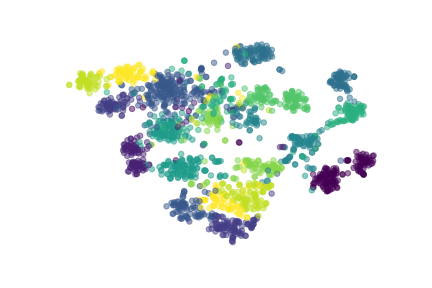}\\
      \vspace{-2.5ex}
      (by class)
      \end{minipage}\\
      \medskip
      \begin{minipage}[t]{0.44\linewidth} \centering \footnotesize
      \includegraphics[width=\linewidth]{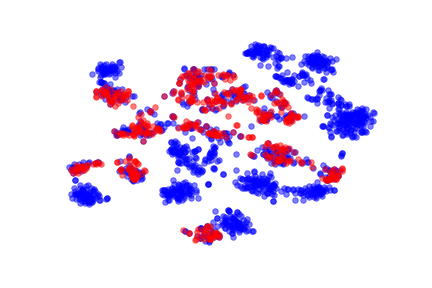}\\
      \vspace{-2.5ex}
      (by domain)
      \end{minipage}
      \begin{minipage}[t]{0.44\linewidth} \centering \footnotesize
      \includegraphics[width=\linewidth]{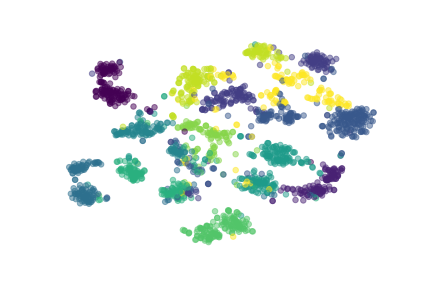}\\
      \vspace{-2.5ex}
      (by class)
      \end{minipage}\\
      \medskip
      (b) VisDA
  \end{minipage}
\caption{
(Best viewed in color)
Additional visualization of feature embeddings on (a) M$\rightarrow$U and (b) VisDA.
Top: source-only, bottom: adapted.
}
\label{fig:feature_tsne_2}
\end{figure*}

\smallskip \noindent
\textbf{Feature Visualization.}
We visualize the feature embeddings for the challenging M$\shortrightarrow$S task
by plotting the outputs of the last pooling layer of the feature generator $G$ using t-SNE \cite{Maaten08}.
In Fig. \ref{fig:feature_tsne}a and Fig. \ref{fig:feature_tsne}b, 
the target (SVHN) features are separated from the source (MINIST) features, forming a large group in the center regardless of their classes. 
However, both domains are better aligned after adaptation, and the features of the same class are in closer proximity in Fig. \ref{fig:feature_tsne}c and Fig. \ref{fig:feature_tsne}d.
More feature embeddings for the tasks of M$\rightarrow$U and VisDA are presented in Fig. \ref{fig:feature_tsne_2}.

%------------------------------------------------------------------------
%------------------------------------------------------------------------

%------------------------------------------------------------------------
%------------------------------------------------------------------------

% -------------------------------------------------------------------
\begin{table}[!t]
\centering
\def\arraystretch{1.0}
\caption{
Ablation study on various source-free UDA tasks.
}
\resizebox{0.98\columnwidth}{!}
{%
\begin{tabular}{l c c c c c c c}
\toprule
~Methods &S$\shortrightarrow$M & M$\shortrightarrow$S & M$\shortrightarrow$U  & U$\shortrightarrow$M  & S$\shortrightarrow$G & O-H & VisDA \\
\midrule
~Source-only
    &70.0 &47.8 &77.4 &84.9 &77.1 &57.9 &56.6 \\
~Only ${\cal L}_{e}$ 
    &99.1 &8.7 &97.7 &95.8 &98.7 &66.7 &78.9 \\
~Only ${\cal L}_{u}$
    &99.4 &42.9 &98.2 &99.0 &5.1 &64.1 &74.0 \\
\midrule
~${\cal L}_{i}$+${\cal L}_f$ ($v$=1) 
    &99.5 &81.7 &98.2 &98.3 &99.4 &71.1 &84.8 \\
~\textbf{FAUST} ($v$=1)
    &{\bf 99.6} &85.9 &98.3 &98.9 &99.6 &71.1 &84.8 \\
~\textbf{FAUST+U} ($v$=1) 
    &{\bf 99.6} &91.1 &{\bf 98.8} &99.0 &{\bf 99.7} &71.7 &84.9 \\
\midrule
~${\cal L}_{i}$+${\cal L}_f$ ($v$=2)
    &99.5 &81.9 &98.1 &98.2 &99.5 &71.6 &{\bf 85.2} \\
~\textbf{FAUST} ($v$=2)
    &{\bf 99.6} &85.9 &98.3 &98.8 &99.5 &71.6 &{\bf 85.2} \\
~\textbf{FAUST+U} ($v$=2) 
    &{\bf 99.6} &{\bf 91.3} &{\bf 98.8} &{\bf 99.1} &{\bf 99.7} &{\bf 72.0} &- \\
\bottomrule
\end{tabular}
}%
\label{tab:ablation}

\end{table}
% -------------------------------------------------------------------

\smallskip \noindent
\textbf{Ablation Study.}
We investigate the effect of each component of our training objective in equation \eqref{eq:total_loss_general}. 
Table \ref{tab:ablation} shows the ablation study on seven source-free UDA tasks. 
When using only ${\cal L}_e$ or ${\cal L}_u$, the results are mostly better than the source-only. 
However, the accuracy collapses on challenging tasks such as M$\shortrightarrow$S and S$\shortrightarrow$G.
On the other hand, the aleatoric uncertainty loss (${\cal L}_i+{\cal L}_f$) consistently shows improved accuracy on various tasks. 
The improvement is more conspicuous on M$\shortrightarrow$S, Office-Home (O-H) and VisDA.
This observation confirms that modeling aleatoric uncertainty plays a significant role in improving the UDA performance.
By constraining the intra- and inter-consistency, the extracted features fall far from the decision boundaries.
Adding ${\cal L}_e$ (FAUST) and ${\cal L}_u$ (FAUST+U) further boost the accuracy.

% -------------------------------------------------------------------
\begin{table}[!t]
\centering
\def\arraystretch{1.0}
\caption{
Effects of the number of views $v$ on Digit and Sign.
}
\resizebox{0.88\columnwidth}{!}
{%
\begin{tabular}{l c c c c c }
\toprule
~Methods &S$\shortrightarrow$M & M$\shortrightarrow$S & M$\shortrightarrow$U  & U$\shortrightarrow$M  & S$\shortrightarrow$G \\
\midrule
~\textbf{FAUST} ($v$=1) 
    &{\bf 99.6}$^{0.0}$ &85.9$^{0.1}$ &{\bf 98.3}$^{0.1}$ &{\bf 98.9}$^{0.1}$ &{\bf 99.6}$^{0.0}$ \\
~\textbf{FAUST} ($v$=2) 
    &{\bf 99.6}$^{0.0}$ &85.9$^{0.2}$ &{\bf 98.3}$^{0.1}$ &98.8$^{0.0}$ &99.5$^{0.1}$ \\
~\textbf{FAUST} ($v$=3)
    &99.5$^{0.1}$ &86.0$^{0.2}$ &98.2$^{0.1}$ &98.8$^{0.1}$ &99.4$^{0.0}$ \\
~\textbf{FAUST} ($v$=4)
    &99.5$^{0.1}$ &85.9$^{0.3}$ &98.2$^{0.0}$ &98.8$^{0.1}$ &99.5$^{0.1}$ \\
~\textbf{FAUST} ($v$=5) 
    &99.5$^{0.0}$ &{\bf 86.1}$^{0.2}$ &98.2$^{0.1}$ &98.8$^{0.1}$ &99.4$^{0.0}$ \\
\bottomrule
\end{tabular}
}%
% \vspace{-1.0ex}
\label{tab:effect_v}

\end{table}
% -------------------------------------------------------------------

\smallskip \noindent
\textbf{Number of Views $v$.}
Increasing $v$ from 1 to 2 improves the performance on Office-Home and VisDA as shown in Table \ref{tab:ablation}.
We investigate the effect of more views in Table \ref{tab:effect_v}.
Due to memory constraints, only smaller datasets are analyzed.
On Digit and Sign, we observe no clear difference in the accuracy with larger values of $v$.
Because augmentation transform is applied on the fly, FAUST sees different views of a given image each time. 
Even with a smaller $v$, FAUST effectively considers many different views as the iterations progress. 
Thus, we suggest that $v$=2 is sufficient in practice.

\smallskip \noindent
\textbf{Augmentation Transforms.}
FAUST leverages random augmented images to model aleatoric uncertainty.
In incorporating data augmentation, we can consider a wide range of augmentation methods, from simple translation and horizontal flips to strong augmentations that heavily distort the given images. 
We compare two augmentation schemes: a standard flip-and-shift (weak) transform and the RandAugment (strong) transform.
RandAugment \cite{RandAugment} is a type of strong augmentation that applies a set of randomly selected transforms with random magnitudes.
In Table \ref{tab:augmentation}, the results are comparable between the two augmentation methods on most UDA tasks.
However, weak augmentation breaks down on the black-and-white to color task M$\shortrightarrow$S.
Weak augmentation also shows a noticeable difference on M$\shortrightarrow$U, where the target data size is relatively small.
These observations suggest that strong augmentation produces more stable and consistent adaptation outcomes.

% -------------------------------------------------------------------
\begin{table}[!t]
\centering
\def\arraystretch{1.0}
\caption{
FAUST+U ($v$=1) results with weak/strong augmentation. 
}
\resizebox{0.92\columnwidth}{!}
{%
\begin{tabular}{c c c c c c c c}
\toprule
~Methods &S$\shortrightarrow$M & M$\shortrightarrow$S & M$\shortrightarrow$U  & U$\shortrightarrow$M  & S$\shortrightarrow$G &O-H & VisDA \\
\midrule
~Weak 
    &99.5 &12.9 &98.1 &{\bf 99.1} &{\bf 99.7} &71.5 & 84.8\\
~Strong 
    &{\bf 99.6} &{\bf 91.1} &{\bf 98.8} &99.0 &{\bf 99.7} &{\bf 71.7} & {\bf 84.9}\\
\bottomrule
\end{tabular}
}%
\vspace{-1.0ex}
\label{tab:augmentation}

\end{table}
% -------------------------------------------------------------------

% -------------------------------------------------------------------
\begin{table}[!t]
\centering
\def\arraystretch{1.0}
\caption{
Complete list of hyperparameters ($\alpha, \beta, \gamma$). 
}
\resizebox{0.98\columnwidth}{!}
{%
\begin{tabular}{c| c| c c c c c c c c}
\toprule
$\gamma$ & $\alpha, \beta$ &S$\shortrightarrow$M & M$\shortrightarrow$S & M$\shortrightarrow$U & U$\shortrightarrow$M  & S$\shortrightarrow$G & O-H & VisDA & miniDN \\
\midrule
\multirow{2}{*}{0} & $\alpha$
    &0.8 &0.2 &0.2 &0.2 &0.5 &1.0 &1.0 &1.0 \\
 & $\beta$ 
    &0.2 &0.8 &0.8 &0.8 &0.5 &0.0 &0.0 &0.0 \\
\midrule
\multirow{2}{*}{1} & $\alpha$ 
    &0.8 &0.5 &0.5 &0.2 &0.5 &1.0 &1.0 &1.0 \\
 & $\beta$ 
    &0.2 &0.5 &0.5 &0.8 &0.5 &0.0 &0.0 &0.0 \\
\bottomrule
\end{tabular}
}%
\vspace{-1.0ex}
\label{tab:hyperparam}

\end{table}
% -------------------------------------------------------------------

\smallskip \noindent
\textbf{Hyperparameters ($\alpha, \beta$).}
For hyperparameters ($\alpha$, $\beta$), we tune the pair in a limited search space \{$(1.0, 0.0)$, $(0.8, 0.2)$, $(0.5, 0.5)$, $(0.2, 0.8)$, $(0.0, 1.0)$\}.
A few labeled target samples are used, similarly to \cite{DIRT-T, DTA}.
A complete list of the selected hyperparameters is presented in Table \ref{tab:hyperparam}.

%==========================================================================
\section{Conclusion}
\label{sec:conclusion}

We proposed a novel UDA method that does not require access to source images.
The proposed method leverages aleatoric uncertainty by employing multiple data augmentations and trains the feature generator by encouraging inter-space consistency and intra-space consistency.
The feature generator is promoted to learn consistent feature representations that are away from the decision boundaries of the fixed head classifier, thus mitigating the divergence between the source and target domains.
We also explored the effect of the epistemic uncertainty loss.
Empirical results demonstrate that our approach outperforms current state-of-the-art methods on various challenging UDA benchmark tasks without using source images 
and better generalizes to the perturbed input images during inference in the target domain.

% ----

% \appendix


%{\small 
\bibliographystyle{elsarticle-num} 
\bibliography{faust}

\begin{thebibliography}{10}
\expandafter\ifx\csname url\endcsname\relax
  \def\url#1{\texttt{#1}}\fi
\expandafter\ifx\csname urlprefix\endcsname\relax\def\urlprefix{URL }\fi
\expandafter\ifx\csname href\endcsname\relax
  \def\href#1#2{#2} \def\path#1{#1}\fi

\bibitem{DAN}
M.~Long, Y.~Cao, J.~Wang, M.~I. Jordan, Learning transferable features with
  deep adaptation networks, in: Proceedings of International Conference on
  Machine Learning (ICML), 2015.

\bibitem{DANN}
Y.~Ganin, E.~Ustinova, H.~Ajakan, P.~Germain, H.~Larochelle, F.~Laviolette,
  M.~Marchand, V.~Lempitsky, Domain adversarial training of neural networks,
  Journal of Machine Learning Research 17(59) (2016) 1--35.

\bibitem{ADDA}
E.~Tzeng, J.~Hoffman, K.~Saenko, T.~Darrell, Adversarial discriminative domain
  adaptation, in: Proceedings of IEEE/CVF Conference on Computer Vision and
  Pattern Recognition (CVPR), Vol.~1, 2017, p.~4.

\bibitem{SAFN}
R.~Xu, G.~Li, J.~Yang, L.~Lin, Larger norm more transferable: An adaptive
  feature norm approach for unsupervised domain adaptation, in: Proceedings of
  IEEE/CVF International Conference on Computer Vision (ICCV), 2019.

\bibitem{MDD}
Y.~Zhang, T.~Liu, M.~Long, M.~Jordan, Bridging theory and algorithm for domain
  adaptation, in: Proceedings of International Conference on Machine Learning
  (ICML), 2019, pp. 7404--7413.

\bibitem{SWD}
C.-Y. Lee, T.~Batra, M.~H. Baig, D.~Ulbricht, Sliced wasserstein discrepancy
  for unsupervised domain adaptation, in: Proceedings of IEEE/CVF Conference on
  Computer Vision and Pattern Recognition (CVPR), 2019.

\bibitem{ImageNet}
J.~Deng, W.~Dong, R.~Socher, L.-J. Li, K.~Li, L.~Fei-Fei, A large-scale
  hierarchical image database, in: Proceedings of IEEE/CVF Conference on
  Computer Vision and Pattern Recognition (CVPR), 2009.

\bibitem{3C-GAN}
R.~Li, Q.~Jiao, W.~Cao, H.-S. Wong, S.~Wu, Model adaptation: Unsupervised
  domain adaptation without source data, in: Proceedings of IEEE/CVF Conference
  on Computer Vision and Pattern Recognition (CVPR), 2020.

\bibitem{SHOT}
J.~Liang, D.~Hu, J.~Feng, Do we really need to access the source data? {S}ource
  hypothesis transfer for unsupervised domain adaptation, in: Proceedings of
  International Conference on Machine Learning (ICML), 2020, pp. 6028--6039.

\bibitem{SDDA}
V.~K. Kurmi, V.~K. Subramanian, V.~P. Namboodiri, Domain impression: A source
  data free domain adaptation method, in: Proceedings of Winter Conference on
  Applications of Computer Vision (WACV), 2021, pp. 615--625.

\bibitem{SoFA}
H.-W. Yeh, B.~Yang, P.~C. Yuen, T.~Harada, Sofa: Source-data-free feature
  alignment for unsupervised domain adaptation, in: Proceedings of Winter
  Conference on Applications of Computer Vision (WACV), 2021, pp. 474--483.

\bibitem{SFDA}
Y.~Kim, D.~Cho, K.~Han, P.~Panda, S.~Hong, Domain adaptation without source
  data, IEEE Transactions on Artificial Intelligence 2~(6) (2021) 508--518.
\newblock \href {https://doi.org/10.1109/TAI.2021.3110179}
  {\path{doi:10.1109/TAI.2021.3110179}}.

\bibitem{AAA}
J.~Li, Z.~Du, L.~Zhu, Z.~Ding, K.~Lu, H.~T. Shen, Divergence-agnostic
  unsupervised domain adaptation by adversarial attacks, IEEE Transactions on
  Pattern Analysis and Machine Intelligence 44~(11) (2022) 8196--8211.
\newblock \href {https://doi.org/10.1109/TPAMI.2021.3109287}
  {\path{doi:10.1109/TPAMI.2021.3109287}}.

\bibitem{ISFDA}
X.~Li, J.~Li, L.~Zhu, G.~Wang, Z.~Huang,
  \href{https://doi.org/10.1145/3474085.3475487}{Imbalanced source-free domain
  adaptation}, in: Proceedings of the 29th ACM International Conference on
  Multimedia, MM '21, Association for Computing Machinery, New York, NY, USA,
  2021, p. 3330–3339.
\newblock \href {https://doi.org/10.1145/3474085.3475487}
  {\path{doi:10.1145/3474085.3475487}}.
\newline\urlprefix\url{https://doi.org/10.1145/3474085.3475487}

\bibitem{Aleatory}
A.~D. Kiureghian, O.~Ditlevsen, Aleatory or epistemic? {D}oes it matter?,
  Structural Safety 31~(2) (2009) 105--112, risk Acceptance and Risk
  Communication.
\newblock \href
  {https://doi.org/https://doi.org/10.1016/j.strusafe.2008.06.020}
  {\path{doi:https://doi.org/10.1016/j.strusafe.2008.06.020}}.

\bibitem{DeepCluster}
M.~Caron, P.~Bojanowski, A.~Joulin, M.~Douze, Deep clustering for unsupervised
  learning of visual features, in: Proceedings of European Conference on
  Computer Vision (ECCV), 2018.

\bibitem{SwAV}
M.~Caron, I.~Misra, J.~Mairal, P.~Goyal, P.~Bojanowski, A.~Joulin, Unsupervised
  learning of visual features by contrasting cluster assignments, in: Advances
  in Neural Information Processing System (NeurIPS), 2020.

\bibitem{MUDA}
J.~{Lee}, G.~{Lee}, Model uncertainty for unsupervised domain adaptation, in:
  Proceedings of IEEE International Conference on Image Processing (ICIP),
  2020, pp. 1841--1845.
\newblock \href {https://doi.org/10.1109/ICIP40778.2020.9190738}
  {\path{doi:10.1109/ICIP40778.2020.9190738}}.

\bibitem{DTA}
S.~Lee, D.~Kim, N.~Kim, S.-G. Jeong, Drop to adapt: Learning discriminative
  features for unsupervised domain adaptation, in: Proceedings of IEEE/CVF
  International Conference on Computer Vision (ICCV), 2019.

\bibitem{CADA}
H.~Zou, Y.~Zhou, J.~Yang, H.~Liu, H.~P. Das, C.~J. Spanos, Consensus
  adversarial domain adaptation, in: Proceedings of the AAAI Conference on
  Artificial Intelligence (AAAI), 2019.

\bibitem{MCD}
K.~Saito, K.~Watanabe, Y.~Ushiku, T.~Harada, Maximum classifier discrepancy for
  unsupervised domain adaptation, in: Proceedings of IEEE/CVF Conference on
  Computer Vision and Pattern Recognition (CVPR), 2018.

\bibitem{DIRT-T}
R.~Shu, H.~H. Bui, H.~Narui, S.~Ermon, A {DIRT-T} approach to unsupervised
  domain adaptation, in: Proceedings of International Conference on Learning
  Representations (ICLR), 2018.

\bibitem{GPDA}
M.~Kim, P.~Sahu, B.~Gholami, V.~Pavlovic, Unsupervised visual domain
  adaptation: A deep max-margin gaussian process approach, in: Proceedings of
  IEEE/CVF Conference on Computer Vision and Pattern Recognition (CVPR), 2019.

\bibitem{STAR}
Z.~Lu, Y.~Yang, X.~Zhu, C.~Liu, Y.-Z. Song, T.~Xiang, Stochastic classifiers
  for unsupervised domain adaptation, in: Proceedings of IEEE/CVF Conference on
  Computer Vision and Pattern Recognition (CVPR), 2020.

\bibitem{DeepJDOT}
B.~B. Damodaran, B.~Kellenberger, R.~Flamary, D.~Tuia, N.~Courty, Deepjdot:
  Deep joint distribution optimal transport for unsupervised domain adaptation,
  in: Proceedings of European Conference on Computer Vision (ECCV), 2018.

\bibitem{RWOT}
R.~Xu, P.~Liu, L.~Wang, C.~Chen, J.~Wang, Reliable weighted optimal transport
  for unsupervised domain adaptation, in: Proceedings of IEEE/CVF Conference on
  Computer Vision and Pattern Recognition (CVPR), 2020.

\bibitem{TSA}
S.~Li, M.~Xie, K.~Gong, C.~H. Liu, Y.~Wang, W.~Li, Transferable semantic
  augmentation for domain adaptation, in: Proceedings of IEEE/CVF Conference on
  Computer Vision and Pattern Recognition (CVPR), 2021, pp. 11516--11525.

\bibitem{FDAN}
J.~Li, M.~Jing, H.~Su, K.~Lu, L.~Zhu, H.~T. Shen, Faster domain adaptation
  networks, IEEE Transactions on Knowledge and Data Engineering 34~(12) (2022)
  5770--5783.
\newblock \href {https://doi.org/10.1109/TKDE.2021.3060473}
  {\path{doi:10.1109/TKDE.2021.3060473}}.

\bibitem{Color}
L.~Bertinetto, J.~Henriques, P.~Torr, A.~Vedaldi, Meta-learning with
  differentiable closed-form solvers, in: Proceedings of European Conference on
  Computer Vision (ECCV), 2018.

\bibitem{Jigsaw}
M.~Noroozi, P.~Favaro, Unsupervised learning of visual representations by
  solving jigsaw puzzles, in: Proceedings of European Conference on Computer
  Vision (ECCV), 2016.

\bibitem{Rotation}
S.~Gidaris, P.~Singh, N.~Komodakis, Unsupervised representation learning by
  predicting image rotations, in: Proceedings of International Conference on
  Learning Representations (ICLR), 2018.

\bibitem{SimCLR}
T.~Chen, S.~Kornblith, M.~Norouzi, G.~Hinton, A simple framework for
  contrastive learning of visual representations, in: Proceedings of
  International Conference on Machine Learning (ICML), 2020.

\bibitem{MoCo}
K.~He, H.~Fan, Y.~Wu, S.~Xie, R.~Girshick, Momentum contrast for unsupervised
  visual representation learning, in: Proceedings of IEEE/CVF Conference on
  Computer Vision and Pattern Recognition (CVPR), 2020.

\bibitem{BYOL}
J.-B. Grill, F.~Strub, F.~Altché, C.~Tallec, P.~H. Richemond, E.~Buchatskaya,
  C.~Doersch, B.~A. Pires, Z.~D. Guo, M.~G. Azar, B.~Piot, K.~Kavukcuoglu,
  R.~Munos, M.~Valko, Bootstrap your own latent: A new approach to
  self-supervised learning, in: Advances in Neural Information Processing
  System (NeurIPS), 2020.

\bibitem{image_aug}
C.~Shorten, T.~M. Khoshgoftaar, A survey on image data augmentation for deep
  learning, Journal of Big Data 6~(60) (2019) 2196--1115.
\newblock \href {https://doi.org/10.1186/s40537-019-0197-0}
  {\path{doi:10.1186/s40537-019-0197-0}}.

\bibitem{AutoAugment}
E.~D. Cubuk, B.~Zoph, D.~Mane, V.~Vasudevan, Q.~V. Le, Autoaugment: Learning
  augmentation strategies from data, in: Proceedings of the IEEE/CVF Conference
  on Computer Vision and Pattern Recognition (CVPR), 2019.

\bibitem{RandAugment}
E.~D. Cubuk, B.~Zoph, J.~Shlens, Q.~Le, Randaugment: Practical automated data
  augmentation with a reduced search space, in: H.~Larochelle, M.~Ranzato,
  R.~Hadsell, M.~Balcan, H.~Lin (Eds.), Advances in Neural Information
  Processing Systems, Vol.~33, Curran Associates, Inc., 2020, pp. 18613--18624.

\bibitem{FixMatch}
K.~Sohn, D.~Berthelot, N.~Carlini, Z.~Zhang, H.~Zhang, C.~A. Raffel, E.~D.
  Cubuk, A.~Kurakin, C.-L. Li, Fixmatch: Simplifying semi-supervised learning
  with consistency and confidence, in: H.~Larochelle, M.~Ranzato, R.~Hadsell,
  M.~F. Balcan, H.~Lin (Eds.), Advances in Neural Information Processing System
  (NeurIPS), Vol.~33, Curran Associates, Inc., 2020, pp. 596--608.

\bibitem{MPL}
Q.~Xie, M.-T. Luong, E.~Hovy, Q.~V. Le, Self-training with noisy student
  improves imagenet classification, in: Proceedings of the IEEE/CVF Conference
  on Computer Vision and Pattern Recognition (CVPR), 2020.

\bibitem{Uncertainty}
A.~Kendall, Y.~Gal, What uncertainties do we need in bayesian deep learning for
  computer vision?, in: Advances in Neural Information Processing System
  (NeurIPS), 2017, pp. 5580--5590.

\bibitem{Gal16}
Y.~Gal, Z.~Ghahramani, Dropout as a {Bayesian} approximation: Representing
  model uncertainty in deep learning, in: Proceedings of International
  Conference on Machine Learning (ICML), 2016.

\bibitem{TTA}
M.~S. Ayhan, P.~Berens, Test-time data augmentation for estimation of
  heteroscedastic aleatoric uncertainty in deep neural networks, in:
  International conference on Medical Imaging with Deep Learning, 2018.

\bibitem{Ben10}
S.~Ben-David, J.~Blitzer, K.~Crammer, A.~Kulesza, F.~Pereira, J.~W. Vaughan., A
  theory of learning from different domains, Machine Learning 79(1-2) (2010)
  151--175.

\bibitem{EntMin}
Y.~Grandvalet, Y.~Bengio, Semi-supervised learning by entropy minimization, in:
  L.~Saul, Y.~Weiss, L.~Bottou (Eds.), Advances in Neural Information
  Processing System (NeurIPS), Vol.~17, MIT Press, 2005.

\bibitem{Advent}
T.-H. Vu, H.~Jain, M.~Bucher, M.~Cord, P.~Perez, Advent: Adversarial entropy
  minimization for domain adaptation in semantic segmentation, in: Proceedings
  of the IEEE/CVF Conference on Computer Vision and Pattern Recognition (CVPR),
  2019.

\bibitem{SVHN}
Y.~Netzer, T.Wang, A.~Coates, A.~Bissacco, B.Wu, A.~Y. Ng, Reading digits in
  natural images with unsupervised feature learning, in: Advances in Neural
  Information Processing System (NeurIPS) Workshop, Vol. 2011, 2011, p.~5.

\bibitem{MNIST}
Y.~LeCun, L.~Bottou, Y.~Bengio, P.~Haffner, Gradient based learning applied to
  document recognition, in: Proceedings of the IEEE, Vol. 86(11), 1998, pp.
  2278--2324.

\bibitem{USPS}
J.~J. Hull, A database for handwritten text recognition research, IEEE
  Transactions on Pattern Analysis and Machine Intelligence (1994) 550--554.

\bibitem{SYNSIG}
B.~Moiseev, A.~Konev, A.~Chigorin, A.~Konushin, Evaluation of traffic sign
  recognition methods trained on synthetically generated data, In International
  Conference on Advanced Concepts for Intelligent Vision Systems, Springer
  (2013) 576--583.

\bibitem{GTSRB}
J.~Stallkamp, M.~Schlipsing, J.~Salmen, C.~Igel, The {G}erman traffic sign
  recognition benchmark: {a} multi-class classification competition, In the
  2011 International Joint Conference on Neural Networks (IJCNN), IEEE (2011)
  1453--1460.

\bibitem{Office-Home}
H.~Venkateswara, J.~Eusebio, S.~Chakraborty, S.~Panchanathan, Deep hashing
  network for unsupervised domain adaptation, in: Proceedings of IEEE/CVF
  Conference on Computer Vision and Pattern Recognition (CVPR), 2017, pp.
  5018--5027.

\bibitem{Visda17}
X.~Peng, B.~Usman, N.~Kaushik, J.~Hoffman, D.~Wang, K.~Saenko, Vis{DA}: The
  visual domain adaptation challenge, in: Proceedings of IEEE/CVF Conference on
  Computer Vision and Pattern Recognition (CVPR), 2016.

\bibitem{MSCOCO}
T.-Y. Lin, M.~Maire, S.~Belongie, J.~Hays, P.~Perona, D.~Ramanan, P.~Dollar,
  C.~L. Zitnick, Microsoft coco: Common objects in context, in: Proceedings of
  European Conference on Computer Vision (ECCV), 2014.

\bibitem{DAEL}
K.~Zhou, Y.~Yang, Y.~Qiao, T.~Xiang, Domain adaptive ensemble learning, IEEE
  Transactions on Image Processing 30 (2021) 8008--8018.

\bibitem{M3SDA}
X.~Peng, Q.~Bai, X.~Xia, Z.~Huang, K.~Saenko, B.~Wang, Moment matching for
  multi-source domain adaptation, in: Proceedings of IEEE/CVF International
  Conference on Computer Vision (ICCV), 2019, pp. 1406--1415.

\bibitem{IEDA}
J.~Choi, Y.~Choi, J.~Kim, J.~Chang, I.~Kwon, Y.~Gwon, S.~Min,
  \href{https://ojs.aaai.org/index.php/AAAI/article/view/6692}{Visual domain
  adaptation by consensus-based transfer to intermediate domain}, in: AAAI,
  Vol.~34, 2020, pp. 10655--10662.
\newblock \href {https://doi.org/10.1609/aaai.v34i07.6692}
  {\path{doi:10.1609/aaai.v34i07.6692}}.
\newline\urlprefix\url{https://ojs.aaai.org/index.php/AAAI/article/view/6692}

\bibitem{GVB}
S.~Cui, S.~Wang, J.~Zhuo, C.~Su, Q.~Huang, Q.~Tian, Gradually vanishing bridge
  for adversarial domain adaptation, in: Proceedings of IEEE/CVF Conference on
  Computer Vision and Pattern Recognition (CVPR), 2020.

\bibitem{CDAN}
M.~Long, Z.~Cao, J.~Wang, M.~I. Jordan, Conditional adversarial domain
  adaptation, in: Advances in Neural Information Processing System (NeurIPS),
  2018.

\bibitem{BNM}
S.~Cui, S.~Wang, J.~Zhuo, L.~Li, Q.~Huang, Q.~Tian, Towards discriminability
  and diversity: Batch nuclear-norm maximization under label insufficient
  situations, in: Proceedings of IEEE/CVF Conference on Computer Vision and
  Pattern Recognition (CVPR), 2020.

\bibitem{ResNet}
K.~He, X.~Zhang, S.~Ren, J.~Sun, Deep residual learning for image recognition,
  in: Proceedings of IEEE/CVF Conference on Computer Vision and Pattern
  Recognition (CVPR), 2016.

\bibitem{LabelSmoothing}
R.~M\"{u}ller, S.~Kornblith, G.~E. Hinton, When does label smoothing help?, in:
  H.~Wallach, H.~Larochelle, A.~Beygelzimer, F.~d\textquotesingle
  Alch\'{e}-Buc, E.~Fox, R.~Garnett (Eds.), Advances in Neural Information
  Processing System (NeurIPS), Vol.~32, Curran Associates, Inc., 2019.

\bibitem{Cutout}
T.~DeVries, G.~Taylor, Improved regularization of convolutional neural networks
  with cutout, arXiv:1708.04552 (08 2017).

\bibitem{ADR}
K.~Saito, Y.~Ushiku, T.~Harada, K.~Saenko, Adversarial dropout regularization,
  in: Proceedings of International Conference on Learning Representations
  (ICLR), 2018.

\bibitem{DRCN}
M.~Ghifary, W.~B. Kleijn, M.~Zhang, D.~Balduzzi, W.~Li, Deep
  reconstruction-classification networks for unsupervised domain adaptation,
  in: Proceedings of European Conference on Computer Vision (ECCV), 2016.

\bibitem{DCTN}
R.~Xu, Z.~Chen, W.~Zuo, J.~Yan, L.~Lin, Deep cocktail network: Multi-source
  unsupervised domain adaptation with category shift, in: Proceedings of
  IEEE/CVF Conference on Computer Vision and Pattern Recognition (CVPR), 2018.

\bibitem{MME}
K.~Saito, D.~Kim, S.~Sclaroff, T.~Darrell, K.~Saenko, Semi-supervised domain
  adaptation via minimax entropy, in: Proceedings of IEEE/CVF International
  Conference on Computer Vision (ICCV), 2019.

\bibitem{Maaten08}
L.~v.~d. Maaten, G.~Hinton, Visualizing data using t-sne, Journal of Machine
  Learning Research 9 (2008) 2579--2605.

\end{thebibliography}
%}

\end{document}